# The Next Decade in AI:
## Four Steps Towards Robust Artificial Intelligence

Gary Marcus
Robust AI
17 February 2020

Abstract

*Recent research in artificial intelligence and machine learning has largely emphasized general-purpose learning and ever-larger training sets and more and more compute.*

*In contrast, I propose a hybrid, knowledge-driven, reasoning-based approach, centered around cognitive models, that could provide the substrate for a richer, more robust AI than is currently possible.*

# Table of Contents

> *[the] capacity to be affected by objects, must necessarily precede all intuitions of these objects,.. and so exist[s] in the mind a priori.*
> — Immanuel Kant
>
> *Thought is ... a kind of algebra ...*
> — William James
>
> *You can't model the probability distribution function for the whole world, because the world is too complicated.*
> — Eero Simoncelli

# 1. Towards robust artificial intelligence

Although nobody quite knows what deep learning or AI will evolve into the coming decades, it is worth considering both what has been learned from the last decade, and what should be investigated next, if we are to reach a new level.

Let us call that new level *robust artificial intelligence*: intelligence that, while not necessarily superhuman or self-improving, can be counted on to apply what it knows to a wide range of problems in a *systematic* and *reliable* way, synthesizing knowledge from a variety of sources such that it can reason *flexibly* and *dynamically* about the world, *transferring* what it learns in one context to another, in the way that we would expect of an ordinary adult.

In a certain sense, this is a modest goal, neither as ambitious or as unbounded as "superhuman" or "artificial general intelligence" but perhaps nonetheless an important, hopefully achievable, step along the way—and a vital one, if we are to create artificial intelligence we can *trust,* in our homes, on our roads, in our doctor's offices and hospitals, in our businesses, and in our communities.

Quite simply, if we cannot count on our AI to behave reliably, we should not trust it.[1]

§

One might contrast robust AI with, for example, *narrow intelligence*, systems that perform a single narrow goal extremely well (eg chess playing or identifying dog breeds) but often in ways that are extremely centered around a single task and not robust and *transferable* to even modestly different circumstances (eg to a board of different size, or from one video game to another with the same logic but different characters and settings) without extensive retraining. Such systems often work impressively well when applied to the exact environments on which they are trained, but we often can't count on them if the environment differs, sometimes even in small ways, from the environment on which they are trained. Such systems have been shown to be powerful in the context of games, but have not yet proven adequate in the dynamic, open-ended flux of the real world.

---

[1] Of course, the converse is not true: reliability doesn't guarantee trustworthiness; it's just one prerequisite among many, including values and good engineering practice; see Marcus and Davis (Marcus & Davis, 2019) for further discussion.

One must also contrast robust intelligence with what I will call *pointillistic* intelligence, intelligence that works in many cases but in fails in many other cases, ostensibly quite similar, in somewhat unpredictable fashion. Figure 1 illustrates a visual system that recognizes school buses in general but fails to recognize a school bus tipped over on its side in the context of a snowy road (left), and a reading system (right) that interprets some sentences correctly but fails in the presence of unrelated distractor material.

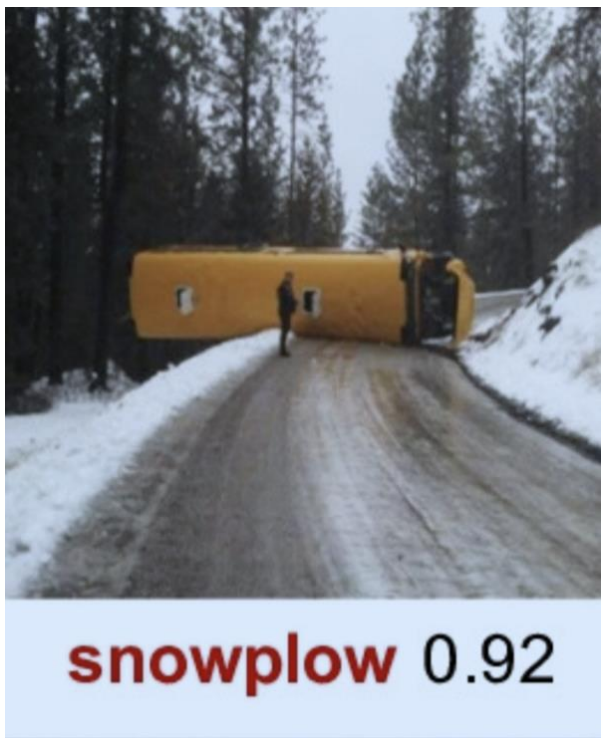

| Sample of how an object in a noncanonical orientation and context fools many current object classification systems (Alcorn et al., 2018) | Sample of how adversarially inserted material fools a large-scale language model (Jia & Liang, 2017) |

*Figure 1: Idiosyncratic failures in vision and language*

Anybody who closely follows the AI literature will realize that robustness has eluded the field since the very beginning. Deep learning has not thus far solved that problem, either, despite the immense resources that have been invested into it.

To the contrary, deep learning techniques thus far have proven to be data hungry, shallow, brittle, and limited in their ability to generalize (Marcus, 2018). Or, as, Francois Chollet (Chollet, 2019) recently put it,

> *AI has … been falling short of its ideal: although we are able to engineer systems that perform extremely well on specific tasks, they have still stark limitations, being brittle, data-hungry, unable to make sense of situations that deviate slightly from their training data or the assumptions of their creators, and unable to repurpose themselves to deal with novel tasks without significant involvement from human researchers.*

In the words of a team of Facebook AI researchers (Nie et al., 2019)

> *"A growing body of evidence shows that state-of-the-art models learn to exploit spurious statistical patterns in datasets... instead of learning meaning in the flexible and generalizable way that humans do."*

A key weakness, as Yoshua Bengio put it in a recent article (Bengio et al., 2019), is that

> *Current machine learning methods seem weak when they are required to generalize beyond the training distribution, which is what is often needed in practice.*

What can we do to take AI to the next level?

§

In my view, we have no hope of achieving robust intelligence without first developing systems with what Ernie Davis and I have called *deep understanding*, which would involve an ability not only to correlate and discern subtle patterns in complex data sets, but also the capacity to look at any scenario and address questions such as a journalist might ask: *who*, *what*, *where*, *why*, *when,* and *how*.

On a good day, a system like the widely discussed neural network GPT-2, which produces stories and the like given sentence fragments, can convey something that ostensibly seems to reflect a deep understanding. Given, for example, a sentence fragment (in bold) like, "Two soldiers walked into a bar", it can often generate a fluent and plausible-sounding continuation that captures, for example, the relation between people, bars, drinks and money:

> **Two soldiers walked into a bar** *in Mosul and spent all of their money on drinks.*

But no matter how compelling many of GPT-2 examples seem, the reality is that its representations are thin and unreliable, akin in to what Nie et al (2019) note above, often falling apart under close inspection (Marcus, 2020). Here are two typical cases, drawn from an in-development benchmark I presented at NeurIPS in December 2019 (Marcus, 2019).

- **Yesterday I dropped my clothes off at the dry cleaners and have yet to pick them up. Where are my clothes?** *at my mom's house.*

- **There are six frogs on a log. Two leave, but three join. The number of frogs on the log is now** *seventeen*

In the first, GPT-2 correctly predicts the category of elements that follows the query fragment (viz. a location) but fails to keep track of where the dry cleaning is. In the second, GPT-2 again correctly predicts the correct response category (in this case a number) and again fails to grasp the detail. As discussed in Marcus (Marcus, 2020; Marcus, 2019) such errors are widespread. We will clearly need a more stable substrate in order to achieve robustness.

§

Business as usual has focused primarily on steadily improving tools for function approximation and composition within the deep learning toolbox, and on gathering larger training sets and scaling to increasingly larger clusters of GPUs and TPUs. One can imagine improving a system like GPT-2 by gathering larger data sets, augmenting those data sets in various ways, and incorporating various kinds of improvements in the underlying architecture. While there is value in such approaches, a more fundamental rethink is required.

Many more drastic approaches might be pursued. Yoshua Bengio, for example, has made a number of sophisticated suggestions for significantly broadening the toolkit of deep learning, including developing techniques for statistically extracting causal relationships through a sensitivity to distributional changes (Bengio et al., 2019) and techniques for automatically extracting modular structure (Goyal et al., 2019), both of which I am quite sympathetic to.

But I don't think they will suffice; stronger medicine may be needed. In particular, the proposal of this paper that we must refocus, working towards developing a framework for building systems that can **routinely acquire, represent, and manipulate abstract knowledge, using that knowledge in the service of building, updating, and reasoning over complex, internal models of the external world**.

§

In some sense what I will be counseling is a return to three concerns of classical artificial intelligence—knowledge, internal models, and reasoning—but with the hope of addressing them in new ways, with a modern palette of techniques.

Each of these concerns was central in classical AI. John McCarthy, for example, noted the value of commonsense knowledge in his pioneering paper "Programs with Common Sense" [McCarthy 1959]; Doug Lenat has made the representation of common-sense knowledge in machine-interpretable form his life's work (Lenat, Prakash, & Shepherd, 1985; Lenat, 2019). The classical AI "blocks world" system SHRLDU, designed by Terry Winograd (mentor to Google founders Larry Page and Sergey Brin) revolved around an internal, updatable cognitive model of the world, that represented the software's understanding of the locations and properties of a set of stacked physical objects (Winograd, 1971). SHRLDU then reasoned over those cognitive models, in order to make inferences about the state of the blocks world as it evolved over time.[2]

---

[2] Other important components included a simple physics, a 2-D renderer, and a custom, domain-specific language parser that could decipher complex sentences like *does the shortest thing the tallest pyramid's support supports support anything green?*

Scan the titles of the latest papers in machine learning, and you will find fewer references to these sorts of ideas. A handful will mention reasoning, another smattering may mention a desire to implement common sense, most will (deliberately) lack anything like rich cognitive models of things like individual people and objects, their properties, and their relationships to one another.

A system like GPT-2, for instance, does what it does, for better and for worse, without any explicit (in the sense of directly represented and readily shared) common sense knowledge, without any explicit reasoning, and without any explicit cognitive models of the world it that tries to discuss.

Many see this lack of laboriously encoded explicit knowledge as advantage. Rather than being anomalous, GPT-2 is characteristic of a current trend away from the concerns of classical AI, and towards a different, more data-driven paradigm that has been powered by the resurgence of deep learning (circa 2012). That trend accelerated with DeepMind's much-heralded Atari game system (Mnih et al., 2015) which, as discussed later, succeeded in playing a wide variety of games without any use of detailed cognitive models.

This trend was recently crystallized in a widely read essay by Rich Sutton, one of founders of reinforcement learning. The essay, called "[The Bitter Lesson](#)", counseled explicitly against leveraging human knowledge:

> *The biggest lesson that can be read from 70 years of AI research is that general methods that leverage computation are ultimately the most effective, and by a large margin…researchers seek to leverage their human knowledge of the domain, but the only thing that matters in the long run is the leveraging of computation. … the human-knowledge approach tends to complicate methods in ways that make them less suited to taking advantage of general methods leveraging computation.*

To some extent, building human knowledge into machine learning systems has even been viewed within machine learning circles as cheating, and certainly not as desirable. In one of DeepMind's most influential paper "Mastering the game of Go without human knowledge", the very goal was to dispense with human knowledge altogether, so as to "learn, *tabula rasa*, superhuman proficiency in challenging domains" (Silver et al., 2017). If common sense could be induced from large-scale corpora, with minimal prior constraint, a large subset of the machine learning community would be immensely pleased.[3] Model-building too, has proven to be hard work, and the general sentiment has been that life would be easier if that step too could be skipped.

§

---

[3] Of course, blindly assimilating all that humans have to say, warts and all, would be problematic in its own way. As ConceptNet's lead maintainer Robyn Speer [put it](#), our ambitions should be better: "We want to avoid letting computers be awful to people just because people are awful to people. We want to provide [knowledge representations] that are not just the technical best, but also morally good."

The problem is, even with massive amounts of data, and new architectures, such as the Transformer (Vaswani et al., 2017), which underlies GPT-2 (Radford et al., 2019), the knowledge gathered by contemporary neural networks remains spotty and pointillistic, arguably useful and certainly impressive, but never reliable (Marcus, 2020).

That spottiness and unreliability is implicit in the kinds of examples above (if you leave your laundry, it obviously can't still be at your mother's house) and in more explicit tests of GPT-2 like these:

> *If you break a glass bottle of water, the water will probably* roll.
>
> *If you break a glass bottle of water, the water will probably* break some more and splatter on the floor. Water creates bubbles, which expand when the amount of water in the bottle increases.
>
> *If you break a glass bottle that holds toy soldiers, the toy soldiers will probably* follow you in there.

Crucially, Sutton's examples for the value of "general methods" in lieu of human knowledge come from *closed-ended* domains, such as games, object classification, and speech recognition, whereas common-sense is open-ended. Winning at a game like Go is very different from interpreting and evaluating a news story or solving an unexpected planning problem in the real world word, like the Apollo 13 situation of figuring [how to solve an air filter issue](#) on an endangered spacecraft where the astronauts are quickly running out of air., a kind of one-off solution that seems well outside the scope of what knowledge-free deep reinforcement learning might manage. When it comes to knowing where the dry cleaning has been left (as in the earlier example, *Yesterday I dropped my clothes off at the dry cleaners and have yet to pick them up. Where are my clothes*), you need an internal model of the world, and a way of updating that model over time, a process some linguists refer to as *discourse update* (Bender & Lascarides, 2019). A system like GPT-2 simply doesn't have that.

When sheer computational power is applied to open-ended domains—such as conversational language understanding and reasoning about the world—things never turn out quite as planned. Results are invariably too pointillistic and spotty to be reliable.

It's time for a rethink: what would our systems look like if we took the lessons of deep learning, but human knowledge and cognitive models were once again a first-class citizen in the quest for AI?

## 2. A hybrid, knowledge-driven, cognitive-model-based approach

Many cognitive scientists, including myself, view cognition in terms of a kind of cycle: organisms (eg humans) take in perceptual information from the outside, they build internal cognitive models based on their perception of that information, and then they make decisions with respect to those cognitive models, which might include information about what sort of entities there are in the external world, what their properties are, and how those entities relate to one another. Cognitive scientists universally recognize that such cognitive models may be incomplete or inaccurate, but also see them as central to how an organism views the world (Gallistel, 1990; Gallistel & King, 2010). Even in imperfect form, cognitive models can serve as a powerful guide to the world; to a great extent the degree to which an organism prospers in the world is a function of how good those internal cognitive models are.

Video games are essentially run according to a similar logic: the system has some kind of internal model of the world, and that model is periodically updated based on user input (and the activities of other entities in the simulated world of the game). The game's internal model might track things like a character's location, the character's health and possessions, and so forth.). What happens in the game (where or not there is a collision after a user moves in particular direction) is function of dynamic updates to that model.

Linguists typically understand language according to a similar cycle: the words in a sentence are parsed into a syntax that maps onto a semantics that specifies things like events that various entities participate in. That semantics is used to dynamically update a model of the world (e.g, the current state and location of various entities). Much (though by no means all) work in robotics operates in a similar way: perceive, update models, make decisions. (Some work, particularly end-to-end deep learning for object grasping does not.)

The strongest, most central claim of the current paper is that if we don't do something analogous to this, we will not succeed in the quest for robust intelligence. If our AI systems do not represent and reason over detailed, structured, internal models of the external world, drawing on substantial knowledge about the world and its dynamics, they will forever resemble GPT-2: they will get some things right, drawing on vast correlative databases, but they won't understand what's going on, and we won't be able to count on them, particularly when real world circumstances deviate from training data, as they so often do.[4]

§

What computational prerequisites would we need in order to have systems that *are* capable of reasoning in a robust fashion about the world? And what it would take to bridge the worlds of deep learning (primarily focused on learning) and classical AI (which was more concerned with knowledge, reasoning, and internal cognitive models)?

---

[4] Would GPT-2 do better if its input were broadened to include perceptual input rather than mere text? Perhaps, but I don't think merely broadening the range of input would solve the system's fundamental lack of articulated internal models. Meanwhile, it is interesting to note that, blind children develop rich internal models and learn quite a bit about language and how to relate it those models, entirely without visual input (Landau, Gleitman, & Landau, 2009).

As a warm-up exercise, consider a simple mission as a stand-in for a larger challenge. Suppose that you are building a machine learning system that must acquire generalizations of broad scope, based on a small amount of data, and that you get a handful of training pairs like these, with both inputs and outputs represented as binary numbers:

| Input | Output |
|-------|--------|
| 0010  | 0010   |
| 1000  | 1000   |
| 1010  | 1010   |
| 0100  | 0100   |

To any human, it quickly becomes evident that there is an overarching generalization (call it a "rule") here that holds broadly, such as the mathematical law of identity in addition, $f(x) = x + 0$. That rule readily generalize to new cases [f(1111)=1111; f(10101)=10101, etc].

Surprisingly, some neural network architectures, such as the multilayer perceptron, described by one recent textbook as the quintessential example of deep learning, have trouble with this. Here's an example of multilayer perceptron, inputs at the bottom, outputs on top, a hidden layer in between; to anyone with any exposure to neural networks, it should seem familiar:

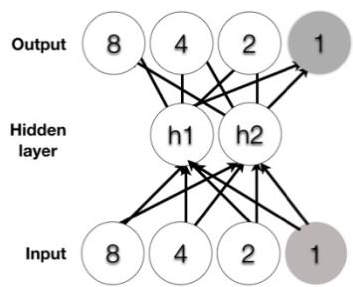

*Multilayer perceptron trained on the identity function*

Such a network can readily learn to associate the inputs to the outputs, and indeed various laws of "universal function approximation" guarantee this. Given enough training data and enough iterations through the training data, the network can easily master the training data.

When all goes well (e.g., if the architecture is set up properly, and there are no local minima in which learning gets stuck), it can also generalize to other examples that are similar in important respects to those that it has seen, to examples that are "within the training distribution", such as these:

| Test Input | Typical Test Output |
|---|---|
| 1110 | 1110 |
| 1100 | 1100 |
| 0110 | 0110 |

But generalizing *outside* the training distribution turns out to be a whole different ballgame:

| Test Input | Typical Human Response | Typical Test Output |
|---|---|---|
| 0011 | 0011 | 0010 |
| 1001 | 1001 | 1000 |
| 1101 | 1101 | 1110 |
| 1111 | 1111 | 1110 |

Such examples show that the multilayer perceptron neural network has *not* after all learned the identity relationship, despite good performance on cases that were within the training distribution. If the same system is trained on f(x)=x for only for even numbers, it will not extend the identity function to odd numbers, which lie outside the training distribution (Marcus, 1998) , To a human, it is obvious from a few examples that each output node, including the rightmost one, which represents the "1" bit ,should be treated in an analogous fashion:  we take the abstraction that we applied the leftmost bit apply it the rightmost digit. A multilayer perceptron trained by backpropagation responds to something different; the rightmost node has always been a zero, and so the network continues to predict that the rightmost node will always be a zero, regardless of the nature of the input, yielding, for example, f(1111)=1110. The network generalizes in its own peculiar way, but it doesn't generalize the identity relationship that would naturally occur to a human.

Adding hidden layers does not change the network's behavior (Marcus, 1998); adding hidden layers with more nodes also doesn't either (Marcus, 1998). Of course any number of solutions can be hacked together to solve the specific problem (learning identity from only even, binary examples), and I have only used the simple identity example here only for expository purposes, but the problem of extrapolating beyond training distributions is widespread, and increasingly recognized.  Joel Grus gives a similar example [here](), with the game fizz-buzz, and Lake and Baroni (Lake & Baroni, 2017) show how some modern natural language systems are vulnerable to similar issues, failing to generalize abstract patterns to novel words in various ways. Bengio made limits on the abilities of extant neural networks central at his recent NeurIPS talk (Bengio, 2019). Within canonical neural network architectures), non-uniform extension of broad universals (such as identity) is surprisingly common, and in my view it remains a central obstacle to progress.

§

In essence, extant neural networks of certain sorts (such as the multilayer perceptrons trained with backpropagation discussed here) excel at two things: memorizing training examples, and interpolating within a cloud of points that surround those examples in some cluster of a hyperdimensional space (which I call generalizing within a training space), but they generalize poorly outside the training space (in Bengio's phrasing, the training distribution).

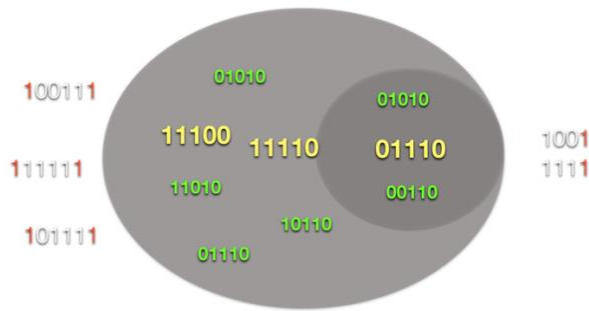

*Multilayer perceptrons: good at generalizing within the space of training examples poor at generalizing the identity function outside the space of training examples.*

What one winds up with in consequence is a pair of closely related problems:

1. **Idiosyncrasy**: systems that lack solid ways of generalizing beyond a space of training examples cannot be trusted in open-ended domains. If you think of each individual system as a function approximator, currently popular systems tend to be great at memorized examples, and good at many (though not all) examples near the training examples—which makes them useful for many applications revolving around classification. But they are poor when pushed beyond the training distribution. A recent math learning system, for example, was good at 1+1=2; 1+1+1=3 up to 1+1+1+1+1=6, but fell apart for 1+1+1+1+1+1+1=7 and all larger examples. (Imagine writing a FOR loop in a computer program, in which one could trust execution only for counter values of less than 7). (By comparison, Microsoft Excel's Flash fill, a symbolic system based on inductive program synthesis, is far more effective in many cases of this variety (Polozov & Gulwani, 2015).

---

5 The exact details of when a system will succeed or not at a given generalization are highly dependent on how problem is set up, e.g., what representational schemes are used; for a fuller discussion, the reader is referred to Marcus, 1998 and Marcus, 2001.

2. **Excessive dependency on exact details of training regime:** whereas all normal human learners acquire their native language and an understanding of the world, despite highly varied circumstances, neural networks often tend to be quite sensitive to exact details such as the order in which training items are presented (hence a literature on "curricula" for neural networks). Likewise, it has been known for three decades that they are vulnerable to a problem known as *catastrophic interference* in which earlier associations are overwritten by later associations (McCloskey & Cohen, 1989), making them highly sensitive to the sequence in which items are presented. Potential solutions continue to be proposed regularly (McClelland, 2019) but the issue remains. Likewise, as one recent paper (Hill et al., 2019) put it, "the degree of generalisation that networks exhibit can depend critically on particulars of the environment in which a given task is instantiated"

§

The idiosyncrasy and inability to extrapolate beyond a training distribution is at odds with the *generality* of much of our commonsense knowledge. It also makes *causality* difficult to reckon with; see also Pearl and Mackenzie (Pearl & Mackenzie, 2018).

Extending an example from the introduction, most ordinary adults and children would recognize (presumably inducing from specific experiences) that the following abstract, *causal* generalization is true:

> IF YOU BREAK A BOTTLE THAT CONTAINS A LIQUID, SOME OF THE LIQUID WILL (OTHER THINGS BEING EQUAL) PROBABLY ESCAPE THE BOTTLE.

Such truths are abstract in that they hold not just for a few specific items but for large, essentially open-ended classes of entities, regardless of what color or shape the bottle or size the bottle is, and whether the bottle contained water, coffee, or an unusual soft drink. We expect a similar generalization to hold for bottles containing ball bearings or game dice, too, even if our prior experience with broken bottles pertained almost exclusively to bottles containing liquids.

Virtually everyone would also recognize that the following generalization as implausible:

> IF YOU BREAK A BOTTLE THAT CONTAINS LIQUIDS, SOME OF THE LIQUID WILL (OTHER THINGS BEING EQUAL) PROBABLY WIND UP 300 METERS AWAY.

Again, regardless of individual experience, we can extend this knowledge in many ways, recognizing that the claim is unlikely to be true of bottles both small and large, even bottles far smaller or larger than those previously encountered.

How can we represent and manipulate and derive knowledge that is that abstract in this sense, pertaining not just to specific entities but to whole classes of things? Challenges in extrapolation mean that common tools like backpropagation-trained multilayer perceptrons on their own are not the right tool for the job. Instead, it is imperative that we find an alternative mechanism for learning, representing, and extending abstract knowledge.

## 2.1. Hybrid architecture

*2.1.1. Symbolic operations over variables offer the only known solution, but the solution is partial*

Symbolic operations over variables offer one potential answer—a solution that is used trillions and trillions of times every second of every day, underlying virtually all of the world's software. In particular, four basic ideas undergird virtually every compute program: variables, instances, bindings that tie variables to instances, and operations over variables.

Each of these ideas is familiar from grade school algebra, in which entities like $x$ and $y$ are *variables*. Specific numbers (2, 3.5, etc) are *instances* that those variables might be *bound* to (e.g, x might currently equal 3). *Operations* include things like addition and multiplication. These make it possible to represent relationships, such as $y = x + 2$, that *automatically* extend to all values in some class (eg all numbers). The process of connecting a variable to instance is sometimes also referred to as *variable binding*.

Computer programs are, of course, built on the same bedrock; algorithms are largely specified in terms of operations that are performed over variables. Variables get bound to instances, algorithms get called, operations are performed, and values are returned.

Importantly, core operations are typically specified in such a way as to apply to all instances of some class (eg *all* integers, all strings, or all floating-point numbers). Core operations typically include basic things like arithmetical operations (addition, multiplication, and so forth), comparisons (is the value of $x$ greater than the value of $y$) and control structures (do something $n$ times, for whatever value the variable $n$ currently happens to be bound to; choose alternative $a$ if the value of $x$ exceeds the value of $y$, otherwise choose alternative $b$, etc). To a first approximation (ignoring bugs, errors in a programmer's logic, etc), this means that properly implemented functions work for all inputs in some class, completely independently of what inputs they may or may not have been exposed to.

It is worth noting that this approach of defining things in terms of functions that are defined in terms of operations is a completely different paradigm from standard machine learning. Whereas machine learning systems typically learn to approximate functions relating input variables to output variables, via a process that Judea Pearl has likened to curve fitting, programmers typically *define* their algorithms independently of training data, in terms of operations over variables. Needless to say, it has served conventional computer programmers well, supporting everything from operating systems to web browsers to video games to spreadsheets, and so on and so forth.

Crucially a system's core operations over variables are generally built to work systematically- *independently of experience*. The mechanics of a [circular bit shift operation](#) in a microprocessor, for example, is defined by a set of parallel suboperations, one for each bit up to the width of the microprocessor's word; the operation works the same whether or not it has ever been used before, and no learning is required. The programmer can safely anticipate that the shift operation will work regardless of experience, and that it will continue in that fashion in the future, again regardless of experience. The virtue of all that machinery—variables, instances, bindings, and operations—is that it allows the programmer to specify things at a certain level of abstraction, with a certain kind of reliability as a byproduct.

Collectively, these four assumptions about variables, bindings, instances, and operations over variables comprise the core of *symbol-manipulation* (Newell, 1980; Marcus, 2001)**.** (Symbols themselves are simply ways of encoding things that get used by other systems, such as a pattern of binary digits used to represent a letter in the ASCII code, or an encoding that allows an output node in a neural network to represent a specific word. So far as I can tell, all current systems make use of them; see Marcus 2001, Chapter 2). Some symbol-manipulation systems might have only a small number of operations, such as addition, concatenation, and comparison, others might have richer operations (e.g., the [unification](#) of complex logical formulae), just as microprocessors can vary in terms of the sizes of their core instruction sets. Recursion can be built on a symbol-manipulating architecture but is not an absolute logical requirement.

As I have argued (Marcus, 2001; Marcus, 1998; Marcus, Vijayan, Bandi Rao, & Vishton, 1999; Berent, Marcus, Shimron, & Gafos, 2002; Berent, Vaknin, & Marcus, 2007) symbol-manipulation in some form seems to be essential for *human* cognition, such as when a child learns an abstract linguistic pattern, or the meaning of a term like *sister* that can be applied in an infinite number of families, or when an adult extends a familiar linguistic pattern in a novel way that extends beyond a training distributions (Berent et al., 2002; Berent et al., 2007). Some of the most compelling evidence for this came from a 1999 study (Marcus et al., 1999) in which my colleagues and I showed that 7-month old infants could recognize simple abstract patterns such as the ABB pattern in *la ta ta* and extrapolate them beyond a set of training examples to novel strings composed entirely of different syllables that didn't phonetically overlap with their training set. Subsequent work shows that even newborns seem capable of this sort of extrapolation. Gallistel and King (Gallistel & King, 2010) have argued that the storage and retrieval of variables is essential for animal cognition. Honeybees, for example, appear to be capable of extending the solar azimuth function to lighting conditions that they have not been exposed to. (Dyer & Dickinson, 1994).

The versatile machinery of symbol-manipulation also provides a basis for structured representations (Marcus, 2001). Computer programs, for example, routinely use tree structures, constructed out of symbols that are combined via operations over variables, to represent a wide variety of things (such as the hierarchical structure folders or directories).

Likewise, the the machinery of symbol-manipulation allows to keep track of properties of individuals as they change over time (e.g., in the form of database records). These capacities too seem central to human language (as in recursive sentence structure) and in our knowledge of individual people and objects as they change over time (Marcus, 2001). (Chapter 5 of *The Algebraic Mind* gives a range of examples that lie outside the scope of eliminative connectionist models, many resting on the persistence of entities over time.)

Such machinery is overwhelmingly powerful. All the world's web browsers, all the world's operating systems, all the world's apps, and so forth are built upon them. (The same tools are also, ironically, used in the specification and execution of virtually all of the world's neural networks).

§

Yet historically mainstream deep learning has largely tried to do without the machinery of symbol-manipulation—often deliberately eschewing it, as a part of a rallying cry for why neural networks offer an alternative to classical paradigms. In the famous PDP books that anticipated much of modern deep learning, Rumelhart and McClelland (, 1986, #39979;) dismissed symbol-manipulation as a marginal phenomenon, "not of the essence human computation". In 2015 Hinton [likened symbols to "luminiferous aether"](#), arguing that the pursuit of symbolic logic as a component of artificial intelligence is "

> *as incorrect as the belief that a lightwave can only travel through space by causing disturbances in the luminiferous aether.... [with] scientists ... misled by compelling but incorrect analogies to the only systems they knew that had the required properties.*

Ideas like database-style records for individuals, too, have been surprisingly absent from the vast preponderance of work on neural networks, and complex structured representations such as hierarchically-structured sentences are found only in a small fraction of research, whereas the norm for both input and output is the simple vector or two-dimensional bitmap, whereas hierarchical data structures and records for individuals studiously avoided.[6]

It doesn't have to be that way. In principle, for example, one could try either to build neural networks that are compatible with symbol manipulation ("implementational connectionism" in terminology introduced by Fodor and Pylyshyn (Fodor & Pylyshyn, 1988) and adopted by Marcus (2001) or neural networks that operate without owing anything thing to the principles of symbol-manipulation ("eliminative connectionism"), or, for some sort of hybrid between the two. The vast majority of work so far has been of the eliminativist variety—but that preponderance reflects sociological fact, not logic necessity.

Within a few years, I predict, many people will wonder why deep learning for so long tried to do so largely without the otherwise spectacularly valuable tools of symbol-manipulation; virtually all great engineering accomplishments of humankind have rested on some sort of symbolic reasoning, and the evidence that humans make use of them in everyday cognition is immense. And in fact, as I will discuss below, things are finally starting to change, with hints of a new, broader pragmatism that I hope will overcome prior dogma.

---

[6] DeepMind's interesting new MEMO architecture (Banino et al., 2020) comes close to representing a database of records.

The first major claim this essay is this: To build a robust, knowledge-driven approach to AI we must have the machinery of symbol-manipulation in our toolkit. Too much of useful knowledge is abstract to make do without tools that represent and manipulate abstraction, and to date, the only machinery that we know of that can manipulate such abstract knowledge reliably is the apparatus of symbol-manipulation.

Alas, by themselves[7], the apparatus of operations over variables tells nothing about *learning*.

It is from there that the basic need for *hybrid* architectures that combine symbol-manipulation with other techniques such as deep learning most fundamentally emerges. Deep learning has raised the bar for learning, particularly from large data sets, symbol manipulation has set the standard for representing and manipulating abstractions. It is clear that we need to bring the two (or something like them[8]) together.

### 2.1.2 Hybrids are often effective

Hybrids are nothing new: Pinker and I proposed three decades ago (Marcus et al., 1992) that the best account of how children learned the English past tense involve a hybrid: a rule (add *-ed* to a verb stem) for forming the past tense of regular verbs, and a neural-network-like system for acquiring and retrieving irregular verbs. And there has long been obvious need for combining symbolic knowledge with perceptual knowledge (e.g., one wants to be able to recognize zebras by combining perceptual knowledge of what horses look like with a verbal definition that likens zebras to horses with stripes[9]).

Computer scientists such as Ron Sun (Sun, 1996) advocated for hybrid models throughout the 1990s; Shavlik (Shavlik, 1994) showed that it was possible to translate a (limited) subset of logic into neural networks. D'Avila Garcez, Lamb, and Gabbay (D'Avila Garcez, Lamb, & Gabbay, 2009) is an important early work on neuro-symbolic approaches. Even Hinton was once warmer to the hybrids, too (Hinton, 1990).

---

[7] Inductive logic programming (Cropper, Morel, & Muggleton, 2019) is a purely-rule based approach to learning that is worth some consideration, though outside the scope of the current paper.

[8] Although I am fairly confident that robust intelligence will depend on some sort of hybrid that combines symbolic operations with machine learning mechanisms, it's not clear whether deep learning (as currently practiced) will play last in its role as dominant machine learning mechanisms, or whether that role will be played some successor that is, e.g., more tractable or more efficient, in terms of data and energy usage. Approaches such as statistical relational learning (Raedt, Kersting, Natarajan, & Poole, 2016) and probabilistic programming (Bingham et al., 2019) that have received much less attention are well worth considering; see van den Broeck (Van den Broeck, 2019) for an overview.

[9] An existing zero-shot learning literature has tried to integrate various forms of multimodal knowledge, but as far as I know, no current system can leverage the precise information that would be found in a dictionary definition.

The bad news is that these early hybrid approaches never got much traction. The results in those days were not compelling (perhaps partly because in those pre-TPU days neural networks themselves then were underpowered). And the neural network community has often been [dismissive](#) of hybrids (and of anything involving symbol-manipulation). All too often, until recently, hybrids have historically been caught in a crossfire between symbolic and neural approaches.

The *good* news is that a long-overdue thaw between the symbol-manipulation world and the deep-learning field seems finally to be coming. Yoshua Bengio, for example, in our December 2019 debate talked about incorporating techniques that could pass variables by name, a standard symbol-manipulating technique used in some earlier computer languages. And there is a growing effort that is actively trying to build symbols and neural networks closer together, sometimes out of practical necessity, sometimes in a research effort to develop new approaches.

Some of the most massive, active commercial AI systems in the world, such as Google Search, are in fact hybrids that mix symbol-manipulation operations with deep learning. While Google Search is hardly what we have in mind for robust artificial intelligence, it is a highly effective AI-powered information retrieval system that works at enormous volume with a high degree of accuracy. Its designers have optimized it extensively in a highly data-driven way and are currently (according to multiple sources) achieving best results by mixing techniques from classic, symbol-manipulating AI (e.g., tools for representing and querying Google Knowledge Graph, which represents knowledge using classic symbolic graph structures that) with tools from the neural network community (e.g., BERT and RankBrain). Google does an enormous empirical experimentation to see what works well on a vast scale, and the fact they still make use of Google Knowledge Graph, even in the era of deep learning, speaks both to the value of symbols and the value of hybrids. (Unfortunately, I know of no detailed public discussion of the relative strengths and weaknesses of the various components.)

AlphaGo is a merger of Monte Carlo Tree Search (wherein a dynamically-constructed, symbolically-represented search tree is traversed and evaluated) and a variety of deep learning modules for estimating the value of various positions. Deep learning alone produces weaker play.

OpenAI's Rubik's solver (OpenAI et al., 2019) is (although it was not pitched as such) is a hybrid of [a symbolic algorithm](#) for solving the cognitive aspects of a Rubik's cube, and deep reinforcement learning for the manual manipulation aspects.

At a somewhat smaller scale Mao et al., (Mao, Gan, Kohli, Tenenbaum, & Wu, 2019) have recently proposed a hybrid neural net-symbolic system for visual question answering called NS-CL (short for the Neuro-Symbolic concept learner) that surpasses the deep learning alternatives they examined. Related work by Janner et al (Janner et al., 2018). pairs explicit records for individual objects with deep learning in order to make predictions and physics-based plans that far surpass a comparable pure black box deep-learning approach. Evans and Grefenstette (Evans & Grefenstette, 2017) showed how a hybrid model could better capture a variety of learning challenges, such as the game fizz-buzz, which defied a multlayer perceptron. A team of people including Smolensky and Schmidhuber have produced better results on a mathematics problem set by combining BERT with a tensor products (Smolensky et al., 2016), a formal system for representing symbolic variables and their bindings (Schlag et al., 2019), creating a new system called TP-Transformer.

Foundational work on neurosymbolic models is (D'Avila Garcez, Lamb, & Gabbay, 2009) which examined the mappings between symbolic systems and neural networks, and showed important limits on the kinds of knowledge that could be represented in conventional neural networks, and demonstrated the value in constructing mixed systems (symbols and neural networks) in terms of representational and inferential capacity. To a first approximation, conventional neural networks can be thought of as engines for propositional logic, and lack good ways of representing quantified statements, as one would find in [predicate calculus with quantifiers](#) such as *every* and *some*). Logic tensor networks (Serafini & Garcez, 2016) aim to implement a formal logic in deep tensor neural networks.

Statistical relational learning (Raedt et al., 2016) represents another interesting approachs that aims to combine logical abstraction and relations with probability and statistics, as does recent work by Vergari et al. on probabilistic circuits (Vergari, Di Mauro, & Van den Broek, 2019). Domingo's Markov Logic Networks seeks to combine symbol-manipulation with the strengths of machine learning (Richardson & Domingos, 2006). Uber's [Pyro](#) (Bingham et al., 2019)

Arabshahi et al., (Arabshahi, Lu, Singh, & Anandkumar, 2019) show how a tree-LSTM can be augmented by an external memory that serves as a stack. Fawzi et al. (Fawzi, Malinowski, Fawzi, & Fawzi, 2019) recently presented a hybrid system for searching proofs of polynomial inequalities. Minervini et al (Minervini, Bošnjak, Rocktäschel, Riedel, & Grefenstette, 2019) recently presented a hybrid neurosymbolic reasoning system called a Greedy Neural Theorem Prover (GNTP) that worked with large-scale databases; Gupta et al (Gupta, Lin, Roth, Singh, & Gardner, 2019) also have made progress in reasoning. The Allen Institute for AI's ARISTO is a complex, multi-part hybrid that has significantly outperformed other systems on eighth-grade science exams (Clark et al., 2019). Battaglia has produced a number of interesting papers on physical reasoning with systems that integrate symbolic graphs and deep learning (e.g., Cranmer, Xu, Battaglia, & Ho, 2019)

And all these are just a few examples of a quickly growing field. It is too early to handicap winners, but there are plenty of first steps towards building architectures that combine the strengths of the symbolic approaches with insights from machine learning, in order to develop better techniques for extracting and generalizing abstract knowledge from large, often noisy data sets.

*2.1.3. Common objections to hybrid models and symbol-manipulation*

Despite the growing interest and multiple considerations in favor of investigating hybrid models, antipathy to symbol-manipulation looms large in some quarters of the machine learning community. As mentioned earlier, Geoffrey Hinton, for example, has argued that European investment into hybrid models would be a "huge mistake", and likened the study of hybrid models to the use of [obsolete gasoline engines in the era of electric cars.](#)

Yet so far as I know Hinton has not written anything lengthy in recent years about *why* he objects to hybrid models that are partly symbolic.

Here are some common objections that I have heard from others, with brief responses to each:

- **Symbols are not biologically plausible**. There are at least four problems with this objection (see also Gallistel and King (Gallistel & King, 2010) for a similar perspective).

    First, just because we have not *yet* decisively identified a neural mechanism supporting symbol-manipulation doesn't mean that we won't ever. Some promising possible neural substrates have already been identified (Frankland & Greene JD, 2019; Marcus, Marblestone, & Dean, 2014; Legenstein, Papadimitriou, Vempala, & Maass, 2016), and other literature has pointed to theoretical plausibly neural substrates (Marcus, 2001). No compelling evidence shows that no such mechanism simply could not exist in the wetware of the brain. Already this year we have seen even a single compartment in a dendrite can compute XOR (Gidon et al., 2020), raising the possibility that individual neurons may be much more sophisticated than is often assumed. The storage and retrieval of values of variables that is central to symbol-manipulation, for example, could act within single neurons (Gallistel & King, 2010).

    Second, a great deal of *psychological* evidence, reviewed above in Section 2.1.1., supports the notion that symbol-manipulation is instantiated in the brain, such as the ability of infants to extend novel abstract patterns to new items, the ability of adults to generalize abstract linguistic patterns to nonnative sounds which they have no direct data for, the ability of bees to generalize the solar azimuth function to lighting conditions they have not directly observed. Human beings can also learn to apply formal logic on externally represented symbols, and to program and debug symbolically represented computers programs, all of which shows that at least in some configurations neural wetware can indeed (to some degree, bounded partly by memory limitations) manipulate symbols. And we can understand language in essentially infinite variety, inferring an endless range of meanings from an endless range of sentences. The kind of free generalization that is the hallmark of operations over variables is widespread, throughout cognition

> Third, the lack of extant *current* evidence of neural realization tells us almost nothing. We *currently* have no detailed understanding of how Garry Kasparov-level chess playing could be implemented in a brain, but that does not mean that Garry Kasparov's chess playing somehow relied on a non-neural mechanism.
>
> Finally, *even if it turned out that brains didn't use symbol-manipulating machinery, there is no principled argument for why AI could not make use of such mechanisms.* Humans don't have floating point arithmetic chips onboard, but that hardly means they should be verboten in AI. Humans clearly have mechanisms for write-once, retrieve immediately short-term memory, a precondition to some forms of variable binding, but we don't know what the relevant mechanism is. That doesn't mean we shouldn't use such a mechanism in our AI.

- **Symbolic systems/hybrid systems haven't worked well in the past**. I have often heard this claimed, but it seems to me to be a bizarre claim. It is simply not an accurate picture of reality to portray hybrid models either as demonstrably ineffective or as old-fashioned, when in fact there is active and effective research into them, described earlier in section 2.1.2.[10]

- **Symbol-manipulation/hybrid systems cannot be scaled**. Although there are real problems to be solved here, and a great deal of effort must go into constraining symbolic search well enough to work in real time for complex problems, Google Knowledge Graph seems to be at least a partial counterexample to this objection, as do large scale recent successes in software and hardware verification. Papers like Minervini et al (Minervini et al., 2019) and Yang et al (Yang, Yang, & Cohen, 2017) have made real progress towards building end-to-end differentiable hybrid neurosymbolic systems that work at scale. Meanwhile. no formal proof of the impossibility of adequate scaling, given appropriate heuristics, exists.

Over the last three decades, I have seen a great deal of bias against symbols, but I have yet to see a compelling argument against them.

*2.1.4. Determining whether a given system is a hybrid system is not always trivial*

A common (though not universal) bias against symbols has given rise to a peculiar sociological fact: researchers occasionally build systems containing the apparatus of symbol-manipulation, without acknowledging (or even considering the fact) that they have done so; I gave some specific examples of this in Marcus, 2001. For example, as noted above the Open AI Rubik's cube solver (OpenAI et al., 2019) contained a [symbolic component](#) known as Kociemba's algorithm, but only a very careful and sophisticated reader would recognize this. The words *hybrid* and *symbolic* were never mentioned. and must be inferred, whereas the word *neural* appears 13 times.

---

[10] It's also worth noting that induction on the past can easily lead to erroneous inferences; deep learning was left for dead in the early 2000's, given results that at the time were not competitive, only to blossom soon thereafter, spurred more by new hardware and larger data sets than fundamental algorithmic change.

Because you can't always tell us how a given system works just from cursory inspection, it's logically possible to unintentionally build a machine that effectively implements symbol-manipulation without any awareness of doing so. Indeed a network designer could stumble onto something that is isomorphic to a symbolic [FPGA](#) without ever knowing it.

While it is certainly conceivable that deep learning systems might offer a genuine alternative to symbol-manipulation, as Bengio suggested in our recent [post-debate dialog](#):

> *My bet is that deep learning variants can achieve the form of symbolic-like computation which humans may actually perform but using a substrate very different from GOFAI, with limitations similar to what humans experience (e.g. only few levels of recursion), and circumventing a major efficiency issue associated with the search problem in GOFAI reasoning in addition to enabling learning and handling of uncertainty.*

we can't take it for granted that any given neural network offers an alternative.

The only way to evaluate whether a system performs an alternative to "symbol-like computation" or computes with bona fide symbol-manipulating operations is to explore *mappings*: to consider that architecture and whether or not its components map onto the components of symbol-manipulation (in something like sense in which chemistry maps onto physics). Marr's (Marr, 1982) levels of computation make it clear that this must be the case: any given computation can be implemented in many ways, and not every implementation is transparent. Chemistry maps onto physics, but that doesn't mean that the mapping was easy to discover. The "right" neural network might or might not map onto symbol-manipulating machinery; the truth may be hard to discern.

My own strong bet is that any robust system will have some sort of mechanism for variable binding, and for performing operations over those variables once bound. But we can't tell unless we look.

§

Lest this sound strange, recall that mapping is no less essential for understanding neuroscience and how it relates to computation. *Whatever computations have been implemented in our brains got there without any conscious decision-making at all; they evolved*. And few of them are transparent. It is the job of neuroscientists and those AI researchers committed to brain-inspired approaches to AI to reverse engineer the brain in order to figure out what computations are there. Whatever drives the brain may or may not map onto our current theories. When we evaluate some theory of how the brain might work, we are evaluating whether the machinery of the brain *does or does not map* onto that theory. Some theories will contain constructs that are isomorphic to actual processes that take place in the brain, others will not. Knudsen and Konishi's (Knudsen & Konishi, 1979) careful work on sound localization in the barn owl is a beautiful example of how one neural circuit was eventually deciphered and mapped onto underlying computations; few research programs have yet equaled it.

Parallel questions arise in AI: when a system works, it is valuable but often nontrivial to understand what drives its performance.

A system that stores every experience in a separate memory than can be retrieved and computed over might be described in "neural' terms yet have components that recognizably play the role of maintaining variables, bindings, instances, and operations (eg retrieval) over variables.

If we create adequate synthetic systems through some sort of search process, be it random, trial-and-error, evolution, AutoML, or other means, we will have solved part of the engineering problem, but *not necessarily yet understood scientifically what makes those models work*. The latter is a job for reverse engineering, and the discovery and rejection of possible mappings, just as it is neuroscience.

If the perfect neural network were to descend on us, we might discover through extensive testing *that* it worked; it would take still another stage of scientific discovery to understand how it worked. If we discover some neural network that succeeds *and it turns out that its constituents should happen to map perfectly onto symbol-manipulation, it will be a victory not only for neural networks but also for symbol-manipulation — regardless of what the system's designers may have intended.* Correspondingly, if none of that system's constituents map onto symbol-manipulation, it would be a defeat for symbol-manipulation.

Any reasonable person will recognize how hard it has been so far to understand how human brains work, and the same will become true for neural networks as they become more complex. The human brain itself is an example of an impressive neural network that has effectively (via evolution) descended upon us; it seems to work quite well, but we have no idea why.[11]

### 2.1.5. Summary

Symbol-manipulation, particularly the machinery of operations over variables, offers a natural though incomplete solution to the challenge of extrapolating beyond a training regime: represent an algorithm in terms of operations over variables, and it will inherently be defined to extend to all instances of some class. It also provides a clear basis for representing structured representations (such as the tree structures that are taken as bedrock in generative linguistics) and records of individuals and their properties.

---

[11] Seeking mappings between implementational details and algorithmic description, if they exist, may also have practical value, since, for example, some low-level neural network-like computations might conceivably be more efficiently computed at a purely symbolic level, once those mappings are discovered. Conversely, some models that are pitched as neural networks, such as Lample and Charton's recent work on symbolic integration (Lample & Charton, 2019) turns out on careful inspection to have serious limits and to be heavily dependent on symbolic processors (Davis, 2019). A clear, principled understanding on how symbolic and neural components work together is likely to be quite valuable.

What it lacks is a satisfactory framework for learning. Hybrids could be a way of combining the best of both worlds: the capacity to learn from large scale data sets, as exemplified by deep learning, with the capacity to represent abstract representations that are syntactic and semantic currency of all the world's computer programming languages. I conjecture that they are a prerequisite for safely arriving at robust intelligence.

Far fewer resources have gone into studying the universe of hybrid models than have gone into "pure" deep learning systems that eschew symbol-manipulating, but the growing strand of work reviewed in section 2.1.2, from a wide range of research labs, to say nothing of the success of Google Search, all point to the value of greater study of hybrid architectures.

Sadly, we are not out of the woods yet, though. Hybrid models that combine powerful data-driven learning techniques with the representational and computational resources of symbol-manipulation may be necessary for robust intelligence, but they are surely not sufficient. In what follows I will describe three further research challenges.

## 2.2. Large-scale knowledge, some of which is abstract and causal

Symbol-manipulation allows for the representation of abstract knowledge, but the classical approach to accumulating and representing abstract knowledge, a field known as knowledge representation, has been brutally hard work, and far from satisfactory. In the history of AI, the single largest effort to create commonsense knowledge in a machine-interpretable form, launched in 1984 by Doug Lenat, is the system known as CYC (Lenat et al., 1985). It has required thousands of person-years, an almost Herculean effort, to capture facts about psychology, politics, economics, biology, and many, many other domains, all in a precise logical form.

Thus far, the payoff has not been compelling. Relatively little has been published about CYC (making evaluation challenging, though see this [interesting Hacker News thread](#)), and the commercial applications seem modest, rather than overwhelming. Most people, if they know CYC at all, regard it as a failure, and few current researchers make extensive use of it. Even fewer seem inclined to try to build competing systems of comparable breadth. (Large-scale databases like Google Knowledge Graph, Freebase and YAGO focus primarily on facts rather than commonsense.)

Given how much effort CYC required, and how little impact it has had on the field as a whole, it's hard not to be excited by Transformers like GPT-2. When they work well, they seem almost magical, as if they automatically and almost effortlessly absorbed large swaths of common-sense knowledge of the world. For good measure,

Transformers give the appearance of seamlessly integrating whatever knowledge they absorb with a seemingly sophisticated understanding of human language.

The contrast is striking. Whereas the knowledge representation community has struggled for decades with precise ways of stating things like the relationship between containers and their contents, and the natural language understanding community has struggled for decades with semantic parsing, Transformers like GPT2 seem as if they cut the Gordian knot—without recourse to any explicit knowledge engineering (or semantic parsing)—whatsoever.

There are, for example, no knowledge-engineered rules within GPT-2, no specification of liquids relative to containers, nor any specification that water even is a liquid. In the examples we saw earlier

> ***If you break a glass bottle of water, the water will probably*** *flow out if it's full, it will make a splashing noise.*

there is no mapping from the concept $H_2O$ to the word *water*, nor any explicit representations of the semantics of a verb, such as *break* and *flow*.

To take another example, GPT-2 appears to encode something about fire, as well:

> ***a good way to light a fire is to use a*** *lighter.*

> ***a good way to light a fire is to use a*** *match*

Compared to Lenat's decades-long project to hand encode human knowledge in machine interpretable form, this appears at first glance to represent both an overnight success and an astonishing savings in labor.

§

The trouble is that GPT-2's solution is just an *approximation* to knowledge, and not substitute for knowledge itself. In particular what it acquires is an approximation to the statistics of how words co-occur with one another in large corpora—rather than a clean representation of concepts per se. To put it in a slogan, it is a model of word usage, not a model of ideas, with the former being used as an approximation to the latter.

Such approximations are something like shadows to a complex three-dimensional world. The concepts of bottles and breaking, via the usage of words associated with them, cast shadows on corpora that encode a subset of human interaction. Transformers analyze the shadows left by the words, like the prisoners in [Plato's allegory of the cave](). The trouble is that that analysis of shadows—the analysis of co-occurrence statistics of how the words are used in those corpora—doesn't necessarily reveal anything about what a bottle actually is or what it is to shatter one.

The truth is that GPT-2 has little clue about what is (and is not) relevant about the *concepts* that underlie the words that it analyzes. Consider what happens for example if, following classic experiments by Frank Keil (Keil, 1992)], you pit perceptual features against deeper conceptual understanding:

> *If you put a costume on a raccoon to make it look a skunk, it will be a* skunk.
>
> *If you paint an airplane so that it looks like a dragon, it will be* a dragon.

Clearly, knowing some statistical properties of how words are used in a concept is not sufficient for understanding the nature of the underlying concepts, such as what makes an airplane an airplane rather a dragon (and vice versa). This recent anonymous, under review manuscript make a similar point.

I am reminded of a Scrabble champions like Nigel Richards, who won a French Scrabble™ tournament without knowing French; the individual words were used as mere game tokens, without any regard whatsoever to their meaning. In essence, GPT-2 predicts the properties of sequences of words as game tokens, without a clue as to the underlying concepts being referenced.

It does this so well that it sometimes induces an "Eliza effect" in which naive humans attribute more to the machine than is really there, but there's never anything more than a superficial conceptual understanding.

We get similar results if we continue to probe the fire-starting example

> *a good way to light a fire is to use* a dry cotton ball to cover the fuse!
>
> *a good way to light a fire is to use* a pint-sized furnace

The words *fire* and *fuse* and *light* and so forth are all intercorrelated; GPT-2 gets that, but that's not enough for GPT-2 to induce a conceptual understanding of the nature of fire; the results are spotty. To put it in an entirely different way, as Judea Pearl might put it, this sort of anomalous behavior is what you get when you try to induce how the world works through approximating the curves of the statistics of how words are used, rather than inducing a more abstract, causal understanding of the world.[12] Sometimes that works, sometimes it doesn't.

One cannot engineer a robust system out of parts with so little guarantee of reliability.

<div style="text-align: center;">§</div>

One problem with trying to build a system out of parts with such little reliability is that downstream inference will inevitably suffer. The whole point of having knowledge is to use it in action and interpretation and decision-making. If you don't know what can cause a fire, or what happens when a bottle breaks, it's hard to make inferences about what is happening around you. Likewise, you cannot reliably plan. Imagine a domestic robot, unable to light a recalcitrant stove, spending hours with dry cotton balls, covering one fuse after the next.

Language understanding inevitably suffers, as well, since we often disambiguate utterances based on their plausibility and context (Tanenhaus, Spivey-Knowlton,

---

[12] There is an active effort within some parts of the deep learning community to try to integrate causal methods; my guess is that this cannot succeed without adding some amount of innate constraint on how causal knowledge is represented and manipulated, likely leading to hybrid networks of some sort.

Eberhard, & Sedivy, 1995). Systems like GPT have some measure of the context of word-usage, but lack reliable representations of cognitive context and plausibility.

Interpretability and explainability would also prove elusive in a system laden with such shallow conceptual understanding. A system that gloms together cotton balls and lighters as equally valid ways of lighting fires may not have internal consistency to satisfy the needs of interpretability.

And where there is no coherent, causal understanding of basic concepts, there may be no way to engineer robustness in complex real-world environments. Pearl is right: if our systems rely on curve-fitting and statistical approximation alone, their inferences will necessarily be shallow.

This brings me to the second major claim of the current paper: *systematic ways to induce, represent, and manipulate large databases of structured, abstract knowledge, often in causal nature, are a prerequisite to robust intelligence*.

## 2.2.1 What kind of knowledge will robust artificial intelligence require?

Here are some basic considerations.

- **Most—but importantly but not all of that knowledge (see below)—is likely to be learned**. No human is born knowing that lighters start fires, nor that dry cotton balls do not, nor what a glass bottle might do when it is broken. One might conceivably hard-wire that knowledge into an AI system, along the lines of CYC manually hard-wiring each fact, but modern enthusiasts of machine learning would obviously prefer not to.  And because there is always new knowledge to be gleaned, mechanisms for learning new abstract, often causal knowledge are a necessity.

- **Some significant fraction of the knowledge that a robust system is likely to draw on is external, cultural knowledge that is symbolically represented**. The great majority of Wikipedia, for example, is represented verbally, and a robust intelligence ought to be able to draw on that sort of knowledge. (Current deep learning systems can only do this to a greatly limited extent.) Much of that knowledge is effectively encoded in terms of *quantified relationships between variables* (e.g., for all *x, y,* and *z*, such that *x, y,* and *z*  are people, person *x* is the grandchild of person *z*  if there is some person *y* that is a parent of *x* and a child of *z*; for all *x* such that *x* is a species, organisms of species *x* give rise to offspring that are also of species *x*, etc).

- **Some significant fraction of the knowledge that a robust system needs is likely to be abstract.** Current systems are good at representing specific facts like BORN(ABRAHAM LINCOLN, KENTUCKY) and CAPITAL(KENTUCKY, FRANKFORT), but lack ways of representing and effectively manipulating information like the fact that *the contents of a bottle can escape, other things being equal, if the bottle is broken*.

- **Rules and exceptions must co-exist**. Regular verbs (*walk-walked*) co-exist with irregular verbs  (*sing-sang*). Penguins that can't fly exist alongside many other birds that can. Machines must be able to represent knowledge in something like the way in which we represent what linguists call generics: knowledge that is generally true, but admits of exceptions (airplanes fly, but we recognize that a particular plane might be grounded) and need not even be statistically true in a proponderance of cases (*mosquitos carry malaria* is important knowledge, but only a small fraction of mosquitos actually carry malaria). Systems that can only acquire rules but not exceptions (e.g. Evans and Grefenstette (Evans & Grefenstette, 2017)]) are an interesting step along the way to building systems that can acquire abstract knowledge, but not sufficient.

- **Some significant fraction of the knowledge that a robust system is likely to be causal, and to support counterfactuals.** Sophisticated people don't, for example, just know that states have capitals, they know that those capitals are determined, politically, by the actions of people, and that those decisions can occasionally be changed. Albany is the present capital of New York State, but if the capital were (counterfactually) burned to the ground, we recognize that the state might choose a new capital. Children know that when glass bottles drop against hard floors, those bottles may break.

- **Although it is relatively easy to scrape the web for factual knowledge, such as capitals and birthplaces, a lot of the abstract knowledge that we have is harder to glean through web-scraping**; few people, for example, write essays about broken bottles and their contents. For the most part, as Lenat once noted, writers don't write down common sense, because their readers already know it. (Blind webscraping has other issues too; for example, historical biases, such doctors being male, tend to get automatically perpetuated by naive scraping systems.)

- **The relevant knowledge needs to be extremely broad in scope.** Understanding a single novel, for example, could require knowledge of technology, political entities, money, weather, human interaction, etiquette, sexuality, violence, avarice, and so forth. The crucial plot turn in John Grisham's first bestselling novel *The Firm*, for example, rested on an understanding of what a photocopier could do, and how fast it could do it, juxtaposed against a deep understanding of human motivations and temporal reasoning.

- **Putting knowledge into practice is hard**. It's one thing to have a giant database of knowledge that includes., e.g, facts about photocopiers and their speed of operation, and another to integrate just that knowledge (amid a vast library of other less relevant information) in the context of mission-critical temporal reasoning concerning the narrow time window in which the heroic lawyer had available before he would be caught in his secretive but noble information gathering act. Connecting abstract knowledge to specific real-world situations at scale in an efficient fashion is essentially an unsolved problem.

- **Some small but important subset of human knowledge is likely to be innate; robust AI, too, should start with some important prior knowledge**. In contemporary ML there is often a strong desire to minimize knowledge and innateness; as discussed below in Section 2.2.3, I think this a mistake.

*2.2.2 Case Study: Containers*

Let's consider a single case study in some detail—the fact that (under ordinary circumstances), *the water in a tea kettle with the lid on can only come out the spout*.

As the reader might expect by now, GPT-2 sort of gets this some of the time. Sort of.

> Completion
>
> **The water in a tea kettle with the lid on can only come out the** top half of the kettle. The hot water that is made from this heat source can flow out the bottom half of the kettle as well. The coffee which is boiled from a kettle with the lid on needs to be left until its finished brewing. If the hot water is allowed to pass under the lid, the steam that is produced is not cold enough and the coffee does not properly heat up, making it unbalanced.

> Completion
>
> **The water in a tea kettle with the lid on can only come out the** top, and there are three possible openings - the tea kettle, the kettle by the pot, and the pot by the lid.

> Completion
>
> **The water in a tea kettle with the lid on can only come out the** top once, with the remainder leaking out underneath. It's the same with a circular

In principle, we might be able to acquire this particular fact through crowdsourcing, but because people rarely state such obvious truths, and even less frequently state them with precision, we shouldn't count on it. And although we might well need to have such a fact in our database, e.g., if we were building the AI to power the decisions of a humanoid eldercare robot, we might not anticipate that need in advance.

It would be better if we could *derive* facts like these, from more general knowledge, such that if we encounter, for example, a tea kettle of unfamiliar appearance we will know what it is and how to interact with it.

Ernest Davis, Noah Frazier-Logue and I, proposed a framework (Davis, Marcus, & Frazier-Logue, 2017) that could help with this sort of challenge: a large set of independently-motivated logical axioms—none specific to tea kettles, all of general utility, largely consisting of abstractions that most ordinary people would on reflection recognize to be true—from which correct inferences about containers might be made.

Overall, the framework in which the axioms are situated is fairly general: axioms about time, space, manipulation, history, action and so forth. The axioms included statements such as the following (a few slightly simplified here for exposition):

- *The physical world consists of a collection of objects, which move around in time over space.*
- *Objects are distinct; that is, one object cannot be part of another or overlap spatially with another.*
- *An object occupies a region of some three-dimensional extent; it cannot be a one-dimensional curve or two-dimensional surface.*
- *A particular quantity of liquid can occupy any region of a specific volume.*
- *A closed container is an object or set of objects that completely envelopes an internal cavity.*
- *An upright open container is an open container with the opening on top.*

An exploratory robot armed with such knowledge (and other machinery for connecting that knowledge with perception and cognitive models) could perhaps then make inferences about the use and function of an unusually shaped kettle with a nearly-hidden spout:

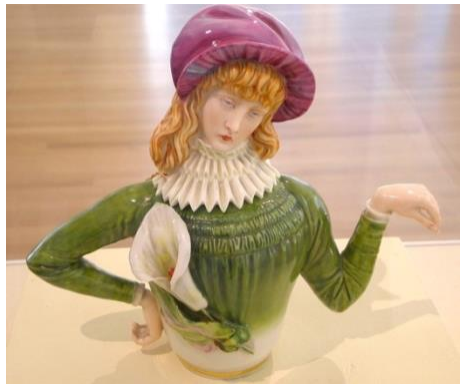

*Tea kettle [believed to have been designed by Richard Williams Binns,](#) (1837-1903) with concealed spout, underneath the hand on the right.*

With some extension, such a system could then provide the foundations of a system that could reason about the affordances of a yarn feeder, even if one had not seen one before; ultimately, one hopes, those foundations could then serve as component in a robotic system that could apply that knowledge in the course of an action, such as knitting.

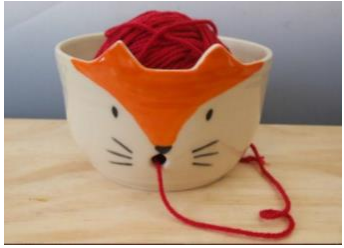

*Yarn feeder. Ball of yarn stays in large opening, and remains there, even as a single strand of yarn is pulled out.*

and draw inferences about (or communicate with a user about) a new feeder, even one that was *grossly* different:

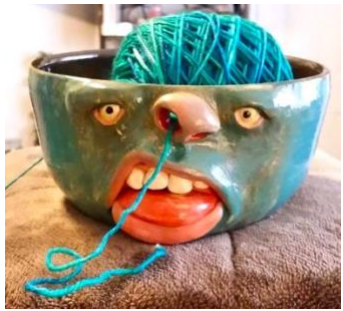

*Yarn feeder 2, pixel-by-pixel rather different.*

This sort of thing is not naturally represented in current neural network approaches. For one thing, doing so would presumably require a prior notion of an object, which itself is not readily represented in most current neural networks. Yet such knowledge needs to be a central part of robust AI, and we must both have strategies for acquiring that knowledge and architectures that can acquire, represent, and manipulate that knowledge -- some of which appears to be symbolic in nature.

In principle, perhaps a smaller subset of the axioms we proposed could be innate, others learned, though as yet I know of no system capable of learning axioms of this sort. (Here again, neurosymbolic systems with embedded knowledge might be helpful.)

Importantly, the set of frameworks themselves cluster into a fairly recognizable set of domains, such as knowledge about space, time, and causality, as sketched here:

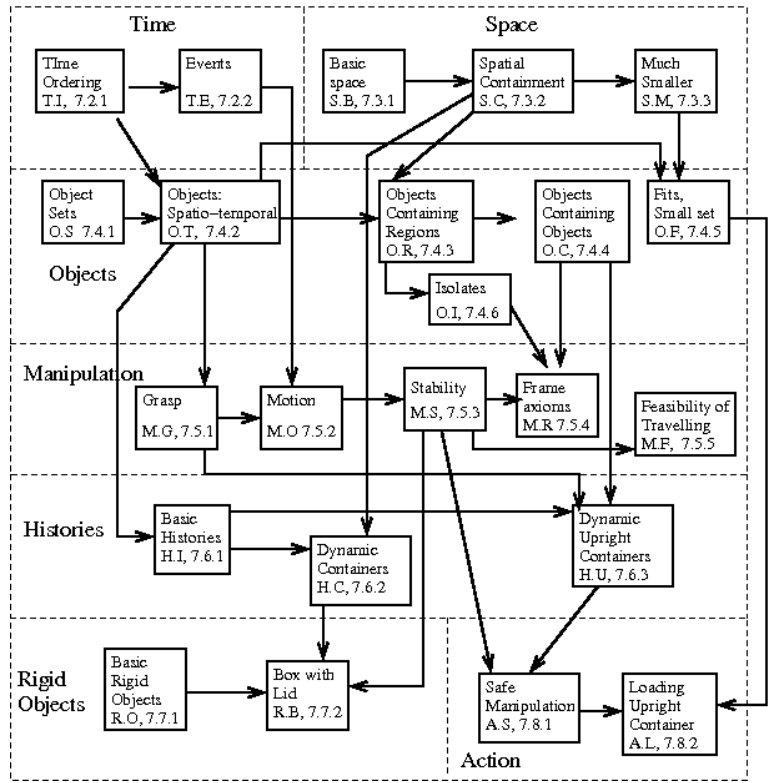

*Framework for knowledge about containers, drawn from* (Davis et al., 2017)

### *2.2.3 Innate Frameworks for Knowledge*

Which brings us to nativism. Although nobody could possibly think that all abstract knowledge is innate, some of it could be, and the argument for having *some* knowledge innate is ultimately quite simple: the more you know at the outset, the easier it becomes to learn the rest; you can constrain the hypothesis space you are trying to search if you can narrow down what you are looking for.

LeCun's seminal work on convolution (LeCun, 1989) was in fact a strong empirical demonstration of precisely this, showing how learning in a digit recognition task, accuracy was better in a system that prewired hierarchical structure equipped with shift-invariant feature detectors (using convolution), relative to a simpler architecture that did not. That single innate prior, convolution, has proven massively valuable.

Yet (many, not all) machine learning researchers resist including *further* innate constraints on their system, drawing what seems to be an arbitrary line, such that network parameters (how many layers, what the loss function is, what input node encoding schemes are used and so forth) are fair game for innateness, but most other things are generally expected to be learned (Marcus, 2020). Some even take this to be a point of pride; as one researcher told me in an in an email, "If you are primarily interested in understanding 'learning' that naturally leads you to disparage 'hand coding.'"

My December 2019 debate with Yoshua Bengio was similarly revealing. He said that it was acceptable to prespecify convolution, because it would only take "[three lines of code](#)", but worried about expanding the set of priors (bits of innate/prior knowledge) far beyond convolution, particularly if those priors would require that more than a few bits of information be specified.

As I expressed to him there, I wouldn't worry so much about the bits. More than 90% of our genome is expressed in the development of the brain (Miller et al., 2014; Bakken et al., 2016), and a significant number of those genes are expressed selectively in particular areas, giving rise to detailed initial structure. And there are many mechanisms by which complex structures are specified using a modest number of genes; in essence the genome is compressed way of building structure, semi-autonomously (Marcus, 2004); there's no reason to think that biological brains are limited to just a few "small" priors. Kevin Mitchell recently summed up the situation nicely, [in a post-debate followup](#).

> *It's certainly true that there isn't enough information in the genome to specify the precise outcome of neural development, in terms of the number and position and connectivity of every neuron in the brain. The genome only encodes a set of mindless biochemical rules that, when played out across the dynamic self-organising system of the developing embryo, lead to an outcome that is within a range of operational parameters defined by natural selection. But there's plenty of scope for those operational parameters to include all kinds of things we would recognise as innate priors. And there is plenty of evidence across many species that many different innate priors are indeed pre-wired into the nervous system based on instructions in the genome.*

And if the *genome* has plenty of scope for innate priors, modern AI systems presumably have scope for even more; we live in an era in which computer memory is measured in gigabytes and terabytes, not bytes or kilobytes. The real question for AI should be not, *how small can we make our library of priors*?, but *what set of priors could most effectively set the stage for learning*? Minimizing the number of bits, per se, really shouldn't be the objective.

§

If there are three proposals for innateness that come up over and over again, they are frameworks for time, space and causality.

Kant, for example, emphasized the value of starting with a "manifold" for time, space, and causality.  Spelke has long argued that some basic, core knowledge of objects, sets and places might be prerequisite for acquiring other knowledge. As she put it (Spelke, 1994),

> *If children are endowed [innately] with abilities to perceive objects, persons, sets, and places, then they may use their perceptual experience to learn about the properties and behaviors of such entities… It is far from clear how children could learn anything about* *the entities in a domain, however, if they could not single out those entities in their surroundings.*

Davis and I have similarly stressed the value of prior frameworks for space, time, and causality both in our analysis of containers (see the figure above (Davis et al., 2017)) and in our recent book (Marcus & Davis, 2019); a host of other developmental psychologists have pointed in similar directions over the years (Landau, Gleitman, & Landau, 2009; Carey, 2009; Leslie, 1982; Mandler, 1992).

As Spelke and Kant both emphasized, once you know about objects and how they travel through time. you can begin to fill in their properties and start to acquire the knowledge that you need for traveling through the world.

The brute fact is that the opposite approach—starting with a near blank slate and training them on massive data sets—simply hasn't panned out so far (Marcus 2020). Relatively blank-slate models trained by large data sets have been given a full-throated test by some of the largest companies in the word (Google, Facebook, Microsoft, and so forth), and provided with almost limitless resources of time, money, computation, and research personnel, yet they still have not been able to reason reliably about time, space, or causality. Absent such capacities, our systems will never be robust enough to cope with the variability of the real-world.

It's surely time to consider a more nativist approach.

§

The good news is that although nativism has often been frowned on in the deep learning community, historically, there are some growing signs of a greater openness to "priors" (e.g., Burgess et al., 2019; Rabinowitz et al., 2018; Maier et al., 2017).

Of course, *every* neural network is actually filled with priors, in the form of innate (i.e., preestablished, rather than learned by the system in question) commitments to particular numbers of layers, particular learning rules, particular connectivity patterns, particular representational schemes (eg what input and output nodes stand for), and so forth. But few of these priors are conceptual in nature.

The real question may be about what *kinds* of priors can be represented, and whether the priors that we need can naturally be represented with the tools we already have, or whether we need new tools in order to represent a broader range of priors. The deep learning community seems to be fine with making use of convolution (which fits naturally within a neural network framework) as a prior, but thus far the community has given far less attention to models with more complex priors, such as innate knowledge about the persistence of objects[13], or a temporal calculus for understanding events over time.

---

[13] See Marcus (2001), Chapter 5 for a discussion of why estimating the trajectory of an object is not in itself sufficient.

What we should be asking is not *what's the least innate structure I can get away with?* but rather *what sorts of priors do I need?*, and *can my existing architectures incorporate them effectively?* Can we build a richer innate basis with a deep learning substrate, or are there limits to what can be conveniently innately represented in that framework? Might we need explicit symbol-manipulating machinery in order to represent other types of abstractions, such as causal ones? Deep learning has proven adept at representing knowledge about what objects look like, but less adept at acquiring and representing knowledge about how the physical world works (Zhang, Wu, Zhang, Freeman, & Tenenbaum, 2016), and about how humans interact with one another, and more generally about causal relationships.

Other approaches, such as probabilistic programming, that allow for explicitly-represented symbolic constraints while at the same time striving to learn from subtle statistical information, are worth serious consideration.

§

Taking a step back to stock, the vast majority of what human beings know about the world is learned:

> *the fact that boats tend to float rather than sink*
>
> *the facts boats do tend to sink if they develop holes below the water line.*
>
> *the fact that lighters are better than cotton balls for starting fires*
>
> *the fact that shattered bottles leak*

and on and on, almost endlessly. Whatever core knowledge we possess clearly must be supplemented by an enormous amount of learned knowledge

It is not unreasonable to think that the average human being might know (or immediately recognize as true) millions or perhaps tens of millions of such facts; the great majority of these must be learned, whether through experience, or explicit instruction, or other means. Importantly, virtually any of that knowledge can be put into practice, governing action and decision-making (e.g., we would choose not to board a boat if we discovered that the boat had a hole below the water line).

But importantly, a significant fraction of our *learned* knowledge is both causal and abstract, which likely necessitates the uses of some sort of hybrid architecture, per the discussion in the previous section.

Meanwhile, pure prewiring will never suffice, because the world itself constantly changes; there will always, for example, be new causal principles associated with new bits of technology. If someone introduces a popular new gadget called a Framulator™, we quickly learn what a Framulator does, how to turn it on and off, and how to make it do its thing. Children do this naturally, as Gopnik and Sobel (Gopnik & Sobel, 2000) have elegantly shown; we need machines that can do the same.

But we probably cannot and should not learn all of our abstract and causal knowledge from experience alone. Doing so would be wildly inefficient, when so much knowledge has already been codified; why, for example, make every system learn anew that objects continue to exist in space and time even when they are occluded, when that is a universal truth? Moreover, as we have seen, eg in the discussion of GPT-2, learning from scratch has thus far been unreliable. Without some prior knowledge, such as the rudiments of physical and psychological reasoning, almost none of what we might call common sense is learned *well*. We need some core knowledge in order to direct the rest of what we learn.

The need for a compromise and innovation is again manifest. We patently need systems (presumably neurosymbolic hybrids) that can acquire new causal knowledge, but in order to acquire that knowledge we *likely need more powerful priors than we have used to date*.

Hence my third major claim:: rather than starting each new AI system from scratch, as a blank slate, with little knowledge of the world, we should seek to build learning systems that start with initial frameworks for domains like time, space, and causalit*y*, in order to speed up learning and massively constrain the hypothesis space.

Whether such frameworks represented in formal logic (a la Cyc) or other means, perhaps not yet invented, I strongly suspect they are a prerequisite for any serious progress towards robust intelligence. No amount of innateness can be a *substitute* for learning, but unfocused learning is not enough. The name of the game is to find the set of innate priors, however small or large, that will best facilitate the learning of the immense library of knowledge our systems ultimately need.

All that said, knowledge by itself is not enough. That knowledge must be put into practice—with tools for reasoning, and in the context of cognitive models—the two topics to which I turn to next.

## 2.3. Reasoning

In a famous anecdote, probably made better in the retelling but [apparently based in a kernel of truth](), the legendary actor Laurence Olivier is on set with the young Dustin Hoffman, who has given up some sleep in order that his character might appear to be exhausted. Olivier says to Hoffman, "Dear boy, you look absolutely *awful. W*hy don't you try acting? It's so much easier."

I feel the same way about memorization versus reasoning. Current approaches to AI are largely trying to cope with the complexity of the world by trying to memorize (or at least approximate) the probability density function for the entire world—at the cost of insatiably needing more and more data. Given the exponential complexity of the world, it's a strategy that is unlikely to work.

Reasoning offers an alternative; instead of memorizing everything, or interpolating between near neighbors that you might have previously encountered, you draw inferences. Instead of memorizing the fact that Plato and Aristotle and Euripides and each of the other billions of other individuals preceding us were all mortal, you learn a general truth, that all humans are mortal, and apply that general truth to specific instances of that category, as needed.

As we have seen, neural nets such as Transformers are (at least as currently typically used, in end-to-end ways, in isolation from the tools of symbol-manipulation) too unreliable to make for sound reasoners. They may work some of the time, but not robustly; symbol-manipulation offers at least the promise of heading the right direction, provided that sufficient knowledge is available.

The best case for reasoning engines in the classical mold comes for the sort of inference that CYC (very much a symbolic system) can perform in optimal circumstances. Take for example, a recent discussion by CYC's founder, Doug Lenat (Lenat, 2019, #3132}, re Romeo and Juliet, abstracted here in two figures that provide a synopsis of the story, some story-relevant knowledge, some common sense knowledge, and an example of the sort of inference CYC (a combination of sophisticated reasoner and large-scale knowledge base) can derive in the best case:

| | | |
|---|---|---|
| "Juliet visits Friar Laurence for help [avoiding being forced by her father to marry Paris], and he offers her a potion that will put her into a deathlike coma for [42] hours. The Friar promises to send a messenger to inform Romeo of the plan so that he can rejoin her when she awakens. …she takes the drug and, when discovered apparently dead, she is laid in the family crypt. "The messenger, however, does not reach Romeo and, instead, Romeo learns of Juliet's apparent death… Heartbroken, Romeo buys poison from an apothecary and goes to the Capulet crypt… Believing Juliet to be dead, he drinks the poison [and dies]. Juliet then awakens and, discovering that Romeo is dead, stabs herself with his dagger and joins him in death." | • T0 – just before Act IV starts.<br>• T1a – the start of Juliet's visit with Friar Lawrence<br>• T1b – the end of that visit<br>• T2 – what the Friar and Juliet both believes about T4 and T5<br>• T3 – Juliet drinks the feign death potion<br>• T3b – Juliet's body is discovered and taken to the crypt<br>• T4 – Romeo hears from the Friar's messenger about the planned trickery<br>• T5 – Romeo hears (from someone else) that Juliet has died<br>• T6 – Romeo goes to the crypt and finds Juliet's inert body.<br>• T7 – Juliet awakens from her feigning of death.<br>• T8 – Romeo and Juliet secretly flee Verona | • Lord Capulet wants and expects Juliet and Paris to marry.<br>• Juliet knows that Lord Capulet wants and expects Juliet and Paris to marry.<br>• Romeo believes that Juliet is alive.<br>• Romeo, Juliet, and the Friar do not want Juliet to marry Paris.<br>• All three of them know that all three of them do not want that to happen. |
| plot summary for Wikipedia | some time points | some statements that are true at $T_0$ |

*Figure 2 Romeo and Juliet, and some samples of story-relevant knowledge, such as specific moments in the story and knowledge at specific time points; from Lenat 2019. As discussed later, the middle and righthand panels reflect part of CYC's cognitive model of the plot.*

| some relevant common-sense knowledge | sample inference |
|---|---|
| • Of course if a person were dead, they would not have to marry anyone.<br>• Of course a person often asks another for help accomplishing something that they both want.<br>• Of course if someone drinks an instantly fatal dose of poison, they immediately die.<br>• Of course if a trusted friend of yours tells you something, and you don't have a better reason *not* to believe it, then you are very likely to believe what they say.<br>• Of course when someone newsworthy is believed to be dead, news of their death will spread quickly. The scale and speed depend on the information technology available and the dead person's level of fame. In particular, in a small medieval European town, the news of a local noble's demise would spread by word of mouth throughout that town over a period of hours but in less than 24 hours. | When she takes the feign-death potion, does Juliet believe Romeo will believe she is alive during the time she is in suspension?<br>○ ✓ Yes.<br>• If, at time T1, an agent's model of a subject's beliefs at time T2 includes a proposition, then the agent believes at T1 that the subject believes the proposition at T2.<br>✓ At the time of Juliet's taking of the feign-death potion, Juliet has a model of Romeo's beliefs at the time of Juliet's being in suspension after taking the feign-death potion that includes the proposition that Juliet is a living thing.<br>• Juliet believes that if an agent receives information from a trusted source, they will believe that information.<br>• Juliet believes that Friar Laurence is a trusted source for Romeo.<br>• Friar Laurence and Juliet plan for Romeo to receive Friar Laurence's message while Juliet is in suspension.<br>• Friar Laurence and Juliet plan for Friar Laurence's message to Romeo to convey the information that while Juliet is in suspension, she will be alive.<br>• Juliet believes the events she and Friar Laurence have planned will occur. |

*FIgure 3 Samples of relevant common sense knowledge, and a complex inference derived by CYC; from Lenat 2019*

The level of detail in the middle and right panels of the upper figure, which list timepoints and statements that hold true at those time points, is an example of what I will call a *rich cognitive model*. It is a *cognitive model* in that it is an internal distillation of a complex scenario, and it is *rich* in that it filled with subtle information about what particular protagonists do and know and intend and expect at particular time points.

We can similarly take the commonsense knowledge illustrated in the left portion of the bottom figure to be quite rich, given the intricacy of the information about actions, consequences, and human interaction that it encodes. Some of these bits of knowledge might be represented explicitly (e.g, the consequence of drinking a certain poison might be instant death), some conceivably might be derived online from more general facts. (For example, "If a person were dead, they would not have to marry anyone" might be derived from a more general observation that obligations tend to apply only to living people.) Either way, the payoff in having such a broad fund of knowledge comes in the sophistication of the inferences that can be drawn (righthand of the bottom figure), and in their level of interpretability—both of which are on a completely different level from anything thus far produced via deep learning.

 Some observations:

- **The approach would not be possible without extensive use of structured representations, operations over variables, and records for individuals**.

- **This represents a best-case proof of concept that shows the potential value of having rich cognitive models and rich knowledge** about biology, theory of mind, and so forth.

- It relies heavily, however, on a great deal of prior work that has done by hand by knowledge engineers that have manually translated Romeo and Juliet into formal logic; **a system that could generate representations of this sort automatically and reason at comparable levels would represent a major breakthrough.**

- **Reasoning per se is not necessarily the bottleneck towards robust AI; the real bottleneck may lie in having the right information accessible in the context of real-world inference.**

Of course, CYC is [far from perfect](). And too little in the world comes in sufficiently prepackaged form for CYC to work with. CYC doesn't have much of a natural language front-end and no vision; to get it to reason, you need to have your problem already represented in logical form. Since very few problems come prepackaged that way, there is comparatively little immediate commercial application. But it is an existence proof that subtle reasoning is possible, within a system that combines large-scale abstract knowledge with higher-order reasoning implemented in a variety of formal logic.

Even in CYC's ability to reason, there is no doubt plenty of room for improvement. Its representations are primarily (or perhaps entirely) the stuff of higher-order logic; it's not clear to what degree it can represent uncertainty, and reason over statistical distributions, and so forth. As Bertrand Russell once put it, "All human knowledge is uncertain, inexact, and partial" and it's not clear how much of that uncertainty, incompleteness and inexactitude CYC can deal with.[14]

And one suspects that, like much else in classical and neural AI, it is likely to be brittle, highly dependent on both the specific knowledge in its databases and the precise ways in which a complex scenario is mapped onto its internal logic.

Still, it–*or something else capable of doing similar work, perhaps using a different approach*– seems like a necessary step along the way to robust reasoning. An AI that can't understand a plot summary of Romeo and Juliet is not likely to be competent in the complexity of the real world. An AI that *can* reason about the motivations of interacting humans as complex events as they unfold over time has a fighting chance.

An optimistic possibility is that reasoning may sort itself out, once the prerequisites of hybrid architecture knowledge are better developed; a pessimistic possibility is that we may need significant improvements in reasoning per se, at the very least in terms of scalability and the ability to deal with incomplete knowledge.  We may not really be able to tell until we get our first two houses--architecture and knowledge representation-in order.

---

[14] Another issue is the many limitations humans face in reasoning, such as confirmation bias, motivated reasoning, context effects, the conjunction fallacy, and so forth. In an ideal world, we would learn from what humans do well, but leave these anomalies behind. For discussions of why humans might have evolved such cognitive inefficiencies even though such cognitive errors may not be functionally optimal see Marcus (Marcus, 2008).

But we can already know this: *because of the complexity of the world, we need something of the sort.* And it is clear that we need new benchmarks to push our systems to do the kind of sophisticated reasoning that the Romeo and Juliet scenario exemplifies. Because we can neither encode every scenario in advance, nor hope always to interpolate between known cases, *a reasoning system that can leverage large-scale background knowledge efficiently, even when available information is incomplete is a prerequisite to robustness*.

Recent work by Minervini et al (Minervini et al., 2019) gives me hope that a neurosymbolic hybrid approach could break new ground. Besold et al. (Besold, Garcez, Stenning, van der Torre, & van Lambalgen, 2017) offer another starting point. The very fact that people are trying gives me more hope; reasoning and knowledge need to be first-class citizens if we are to move forward, and it is good to see people trying.

## 2.4. Cognitive models

A special sort of knowledge is the knowledge that accumulates over time about particular states of affairs, such as what we might learn about a friend in the course of a conversation, about a nation in the course reading the news, or about a set of people as we read a book. In cognitive psychology, we call such accumulating representations, *cognitive models.* Your cognitive model might differ from mine; yours might be more detailed, mine might be less detailed, but we both use them, routinely. But minimally, a cognitive model might consist of knowledge about some set of entities (e.g., the characters in a story and the objects they have available to them), some set of properties (e.g., the size and colors of the objects, the goals of the characters, etc), and information about time and events (at what time *t* did character *x* meet character *y*, and what did *x* know at time *t*).

The propositions and time markers in the CYC/Romeo and Juliet illustration, such as sets of explicit representations of complex facts about believed what and when, are one example of what a rich cognitive model might encode in an AI system. One can also think of Johnson-Laird's (Johnson-Laird, 1983) work on mental models. If I tell you that there is an empty bookshelf, and explain that I then place two books on the shelf, you construct some sort of internal representation of bookshelf containing two books. If I then tell you that I have added another book to the shelf, you *update* your representation such that you now have an internal representation of a bookshelf containing three books. To understand something is, to a large degree, to infer a model of what it is, and ultimately to be able to make inferences about how it operates, and what might happen next.

This is by no means a trivial process. Any GOFAI researcher can describe how, in principle, one can construct (some) complex cognitive models manually, but inferring what the right cognitive model might be on a particular occasion is a complex process, often with more than one plausible answer in a given situation, and in no way yet automated.

The CYC example Romeo and Juliet is compelling in that the inferences the system derives are sophisticated and sensible, but disappointing in that the underlying model is hand-coded, rather than induced from a script of the play. That makes the system fine for demonstration purposes, but in the real world, for cognitive models to pave the way to robust artificial intelligence, we to find ways of inferring them from streams of data (such as video or texts) *automatically*.

That's such a hard problem that (outside of the domain of scene comprehension, discussed below) most people instead work on other things, and to a surprising extent try to make do without cognitive models altogether.

<div style="text-align:center">§</div>

DeepMind's Atari game system, DQN, for example, almost entirely lacks explicit cognitive models. When DQN learned to play Breakout it did not abstract individual board positions into scene graphs representing the location and extent of individual bricks; there was no direct representation of where the paddle is, the velocity of the ball, or the underlying physics of the game, nor any abstract realization of the ricochet dynamic that makes the game so engaging. In the language of reinforcement learning, the system was model-free.[15] Yet superhuman achievement was achieved. (Remarkably, in some games, like Pong, that are strictly deterministic with known starting conditions, successful play can be achieved without looking at the screen at all (Koul, Greydanus, Fern - arXiv preprint arXiv:1811.12530, & 2018, ).

But what is the lesson from the success of systems like DQN? In my opinion, the field has overgeneralized. In closed-end domains like Breakout, model-free reinforcement learning often, given enough data (generally far more than humans would require in similar circumstances), works remarkably well. But that doesn't mean that model-free reinforcement learning is a good *general* solution to intelligence.

The catch is that model-free solutions generalize poorly outside the exact environments in which they were induced. Kansky et al., (Kansky et al., 2017) showed this in a compelling way by tinkering with Breakout; even minor changes—such as moving the paddle up a few pixels up—led to drastic reductions in performance. A human—working with an internal cognitive model— could quickly compensate; model-free deep reinforcement learning systems often can't, instead frequently requiring considerable retraining, precisely because they lack rich cognitive models of the environments in which they operate.[16]

---

[15] Hair-splitters might argue that there is some sort of internal model that is self-generated, pointing to internal states of the system that to some degree correlate with classic cognitive states, the more so with a system like MuZero (Schrittwieser et al., 2019). In my view, the limited ability of such systems to transfer to novel circumstances (see main text) undermine any strong claims of this sort.

The range of failures in language understanding of current Transformers such as GPT-2 (see Marcus, 2019, 2020) reflects something similar: the schism between predicting general tendencies (like the likelihood of the phrase *mom's house* appearing the neighborhood of the words and phrases such as *drop, off, pick, up* and *clothing* in the corpora GPT-2) and the capacity to represent, update, and manipulate cognitive models. When BERT and GPT-2 failed to track where the dry cleaning would be it was *a direct reflection of the fact GPT and BERT have no representation of the properties of individual entities as they evolve over time*. Without cognitive models, systems like these are lost. Sometimes they get lucky from statistics, but lacking cognitive models they have no reliable foundation with which to reason over.

The lack of cognitive models is also bleak news for anyone hoping to use a Transformer as input to a downstream reasoning system. The whole essence of language comprehension is to derive cognitive models from discourse; we can then reason over the models we derive. Transformers, at least in their current form, just don't do that Predicting word classes is impressive, but in of itself prediction does not equal understanding.

§

As we saw with the Romeo and Juliet example in Section 4.3, CYC is immensely better off inasmuch as it *can* (at least to some non-trivial degree) reason over cognitive models (eg its list of time points and facts about characters and locales known at various time points, excerpted in Figure 2) in association with background (common sense knowledge)--but still tragically flawed because it cannot, on its own, *derive* the relevant cognitive models. **Any system that could take natural language Wikipedia plot summaries on their own (such as the one in Figure 2 left panel) as input and automatically *derive* detailed cognitive models on its own (of the sort CYC's programmers manually constructed) such that downstream reasoners could reason over them would be a major step forward relative to current AI**.

Unfortunately, very few people are working on deriving rich cognitive models from texts (let alone videos) that describe events that develop over time A paper by Pasupat and Liang (Pasupat & Liang, 2015) tries to parse sentences into programmable queries that can be run over tables, but that system doesn't try to accumulate models over time. A few papers from Facebook AI Research on models such as Memory Networks (Bordes, Usunier, Chopra, & Weston, 2015) and Recurrent Entity Networks (Henaff, Weston, Szlam, Bordes, & LeCun, 2016) that can take simple stories as input and answer some basic questions about them. But these systems (a) require a large amount of input relative to each question that they answer and (b) seem to be limited in scope, relying to

---

[16] Interesting work towards integrating more Ha and Schmidhuber (Ha & Schmidhuber, 2018) and (Veerapaneni et al., 2019).

a large degree on verbal overlap between question and answer and (c) have very limited ability to incorporate prior knowledge. Perhaps most importantly, (d) they don't produce rich cognitive models that could be passed to reasoners as their output. Peter Norvig's dissertation on story understanding (Norvig, 1986) attempted to do something like this in a classic symbol-manipulation framework, as was much of Schank's and Abelseon's (Schank & Abelson, 1977) seminal work, but so far as I know story understanding is no longer an active current area of research. It is a vital question that has been abandoned, rather than resolved. (Schank & Abelson, 1977)

The closest active literature I know of is work on scene comprehension, which ultimately aims to interpret visual scenes not only in terms of what objects are there, but how the objects relate to one another, e.g. not just identifying a glass and a table, but taking note of the fact that a particular glass is on a table, in a particular room, that that glass is near the edge of the table and that it is supported by the table, and so forth. This is already somewhat beyond the state of the art[17]; ultimately, cognitive model induction needs to go much further; we also need for example to identify psychological relationships, both at a superficial level eg. person 1 is talking to person 2, and ultimately at a more sophisticated level (e.g. person 1 is talking to person 2 in order to deceive person 2, such that person 2 will give person 1 money). At least some of our symbols must in some way ground in our perceptual experience, and if we are to interpret scenes in terms of symbols, we must have ways of inferring symbols (and structured relationships among symbols) from input. Building adequate models will also require systems that can infer temporal boundaries and temporal relationships, and so forth.

Scene comprehension is just one instance of a larger problem; we need to do the same thing every time we understand a story, or read an article, in this case from words rather than direct visual experience.

In our first forays into robust intelligence, we cannot expect to build machines that comprehend Shakespeare, but we can aspire for much more than we've got. My children, age 5.5 and 7, might not spontaneously comprehend Shakespeare, but the intelligence that they do have is, for the most part, robust; they understand much of what there is to know about the physical interactions of everyday objects, and enough about human goals and motives to understand an enormous number of children's books; already, early in elementary school, they have grasped concepts such as deception and misapprehension and motivation that are key in the Romeo and Juliet story. They can climb and maneuver through playgrounds of immense variety, and converse on a vast (though not infinite) range of topics.

---

[17] In my view current work scene comprehension is poor, in part because a lot of the work tries to recognize scenes as a whole ("person cooking meal") rather than in terms of sets of individuals (such as people or objects) and relations between those entities, and in part because it aims to do largely without reasoning, and reasoning is sometimes essential for constructing models that are coherent. Because the number of possibilities is exponentially large, techniques that are suitable for classifying images from finite set of categories unlikely to suffice; rather, reasoning itself must contribute (alongside of object classification) to the process of inducing cognitive models from scenes. One promising recent probabilistic generative model, GENESIS, explicitly models dependencies between scene components (Engelcke, Kosiorek, Jones, & Posner, 2019). See also DeepMind's MONet (Burgess et al., 2019)and MERLIN (Wayne et al., 2018) as well the expressive generative model in (Gregor et al., 2019)] and. e.g., inverse graphics papers by Vicarious (Kansky et al., 2017; George et al., 2017) and Josh Tenenbaum's group (Mao et al., 2019; Veerapaneni et al., 2019).

Ultimately, reasoning and cognitive models can be combined, in an almost infinite number of ways. Suppose, for example, that at 12:00 pm a child has been left alone in a room with a closed cookie jar containing a cookie. At 12:05pm it is observed that the jar is closed, but the cookie is nowhere to be seen. What happened in between? Combining temporal and spatial reasoning, you can readily infer that (a) The child opened the jar (b) the child removed the cookie, (c) the child ate the cookie, and (d) the child closed the jar. For bonus points you can infer that (b) must have happened before both (c) and (d), but also that the order of (c) versus (d) is unknown. In conjunction with a theory of biology, you can construe the child as a container and realize that the cookie is now contained (partly digested) within, having traversed from an aperture (a mouth) into yet another container (the stomach) within. There is no reason whatsoever to think that AI will be able to make these kind of inferences robustly without both internal cognitive models and mechanisms to reason over them. Without this, there is no way to reliably understand a detective story, a conversation that goes beyond small talk, or virtually any narrative of human interaction.

Two conjectures I would make are these

- **We cannot construct rich cognitive models in an adequate, automated way without the triumvirate of hybrid architecture, rich prior knowledge, and sophisticated techniques for reasoning.** To take one example, if we saw ripples in a body of water that were vaguely reminiscent of a car, under ordinary circumstances, we ought to assume that those ripples are just ripples, based on e.g., the knowledge that cars don't float. But we might change our priors in the context of a gangster movie, in which cars might deliberately be pushed into the water. Scene comprehension is ultimately not just about labeling objects, it's about making a coherent interpretation using the best available data, and that requires some degree of reasoning over that data, in conjunction with prior knowledge. Any purely-bottom up approach that relied on labeled images alone is likely to fail in a variety of outlier cases, as we have seen in efforts thus far even in the narrow domain of autonomous driving on well-mapped public roads, despite tremendous efforts and billions of dollars in investment.

- **We cannot achieve robust intelligence without the capacity to induce and represent rich cognitive models.** Reading, for example, can in part be thought a function that takes sentences as input and produces as its output (internal) cognitive models. When we discover that Romeo has killed himself, based on a misapprehension of the facts, we update our model. Once we update our models, we reason over them (does it make sense Romeo would kill himself, given Juliet's apparent death?). Our emotional response, too, is derived from relative judgements about our internal cognitive model of what has happened. (Was the action the character just performed moral? safe? foolish? tragic?). Pure co-occurrence statistics have not reliably gotten to any of this. Cyc has the capacity to represent rich cognitive models, but falls down on the job of inducing models from data, because it has no perceptual component and lacks an adequate natural language front end. Transformers, to the extent that they succeed, skip the steps of inducing and representing rich cognitive models, but do so at their peril, since the reasoning they are able to do is consequently quite limited. Domestic robots, similarly, will ultimately be bound in the range of actions they can safely participate in by their ability to induce, represent, and reason over rich cognitive models.

§

More data alone is probably not going to solve this. As I was editing this manuscript, Google released Meena (Adiwardana et al., 2020), trained on a massive 341 GB corpus, almost ten times the size of what GPT-2 was trained on, and the equivalent of roughly 341,000 books, far more than most people read in a lifetime.

[According to Google's blog](), "Towards a Conversational Agent that Can Chat About…Anything", Meena "can conduct conversations that are more sensible and specific than existing state-of-the-art chatbots." Like GPT-2, Meena produces prose that is both remarkably fluent and also remarkably on topic. Its best dialogs are stunning, and the paper reports that it sets a new state of the art in new measure of "Sensibleness and Specificity". But the same core foundation—an internal cognitive model of the world—is still missing.

Consider this snippet, drawn from Conversation 61 at https://github.com/google-research/google-research/blob/master/meena/meena.txt

> *Human: What's your favorite band?*
>
> *Meena: Avenged sevenfold.*
>
> *Human: Ooh, that's a good answer! What's your least favorite band?*
>
> *Meena: Avenged sevenfold.*
>
> *Human: Oh, i thought you liked Avenged sevenfold?*
>
> *Meena: Yeah. That's my favorite band.*

Each answer is coherent, but there is no coherence from one answer to the next. Meena captures the fact that certain phrases like band names are often correlated with phrases like "favorite band"; but there is no enduring record of Meena's preferences, even for the length of a short conversation, and hence complete inconsistency.

Much the same can be seen in Conversation 77. Meena wavers between describing itself as "learning" Cantonese, and already being natively fluent in Cantonese:

*Human: What are some of your new year resolutions?*

*Meena: Lose 60 pounds Learn Cantonese Learn French*

*Human: Don't you think, that's pretty far fetched? :P*

*Meena: Not really. I'm currently down 35 pounds, I can already speak Cantonese to a basic level, and I'm already learning French in school.*

*Human: Wao! That's impressive!*

*Meena: I'm learning French on the side, it's really easy for me since I'm a native Cantonese speaker.*

There is no recognition that you can't learn a language that you already know (missing background knowledge) and no underlying internal model of whether Meena does or does not speak Cantonese. The sentence-by-sentence predictions are outstanding; the totality is incoherent.

§

While I was revising an early draft of this paper I started drafting a figure with the sentence The cat is on the mat. My plan was to show how the sentence was interpreted, following a path from syntax to semantics to updatable cognitive models that would keep track of entities (e.g, cats and mats) and their properties and relations between one another; the goal was to show how GPT-2 was trying to short-circuit that path, to mixed results.

 Before I could finish drafting the figure, though, my 5.5- and 7-year-old children looked over my shoulder, and read the cat on mat sentence aloud, giggling.  I turned to the older one and asked him, "could you put an elephant on the mat?" He answered, it depends; if it was a really big mat, you could, if it was a little mat you couldn't. He had instantaneously formed a model of a fictional world and the entities that populated that world, and applied his general commonsense knowledge to reason about the world, entirely without labeled examples.

When he left the room, I quizzed his sister, my 5.5 year-old daughter. She understood the earlier conversation perfectly well, and provided a similarly appropriate, it-depends answer to my elephant and mat query.  When I then asked her whether a house could fit on the mat, she proved equally adept at constructing a model and reasoning over its unspecified parameters in order to derive reasonable conclusions.

There is just no way we can build reliable, robust AI with systems that cannot match the basic reasoning and model construction young children routinely do. Waiting for cognitive models and reasoning to magically emerge from larger and larger training corpora is like waiting for a miracle.

The bottom line is this: too little current research is being directed towards building systems with cognitive models. The emphasis on end-to-end learning with massive training sets has distracted from the core of what higher-level cognition if about. Most researchers aren't even trying to build systems that revolve around cognitive models, and (except in narrow domains like autonomous driving) ever fewer are focusing on the related challenge *of* discovering general ways of *deriving and updating cognitive models relative to streams of input* (such as text or video). Even fewer are focused on reasoning about such models in conjunction with prior commonsense knowledge, such as the size of an elephant relative to a cat, and how that relates to mats of various sizes.

In my view, building systems that can map language and perceptual input into rich, evolving cognitive models should be one of the highest priorities in the field.

To put it somewhat differently, and more urgently, every moment spent on improving massive models of word-level prediction, a la GPT-2 and Meena, is (despite potential short-term utility. e.g., in improving translations) a moment that might be better spent on developing techniques for deriving, updating, and reasoning over cognitive models.

If we want to build robust AI, we can ill afford to wait.

# 3. Discussion

## 3.1. Towards an intelligence framed around enduring, abstract knowledge

Without us, or other creatures like us, the world would continue to exist, but it would not be described, distilled, or understood. A bird might flap its wings, and the bird might be carried along in flight. There would be correlation, but not causal description. Human lives are filled with abstraction and causal description. Our children spend a large part of their time asking why; scientists ask such questions in order to generate theories. A significant part of our power comes from our effort to understand and characterize the world, in the form of science, culture, and technology.

Much of that effort culminates in the form of knowledge, some specific, some general, some made verbal, some not. A large part of the goal of classical AI was to distill such knowledge in machine-interpretable form; CYC was the largest project in that vein.

Somewhere along the way, the field of AI took a different direction. Most researchers, if they know CYC at all, regard it as a failure, and few current researchers would describe their goal as accumulating knowledge in anything like the sense that Lenat described.[18]

The partial success of systems like Transformers has led to an illusory feeling that CYC-scale machine-interpretable representations of human knowledge is unnecessary, but I have argued that this is a mistake. As we have seen, however, although Transformers are immensely impressive as statistical inference engines, they are a long way from being a sound basis for robust intelligence. They are unreliable, their knowledge spotty.

---

[18] Perhaps Google Knowledge Graph comes closest, but from what I understand, the goal of Knowledge Graph is to accumulate specific facts that can help in disambiguating search queries, such as the fact that there is a city called Paris in France, rather than abstract common sense.

They reason poorly, and they fail to build cognitive models of events as those events unfold over time; there is no obvious way to connect them to more sophisticated systems for reasoning and cognitive model building, nor to use them as framework for interpretable, debuggable intelligence.

The burden of this paper has been to argue for a shift in research priorities, towards four cognitive prerequisites for building robust artificial intelligence: hybrid architectures that combine large-scale learning with the representational and computational powers of symbol-manipulation, large-scale knowledge bases—likely leveraging innate frameworks—that incorporate symbolic knowledge along with other forms of knowledge, reasoning mechanisms capable of leveraging those knowledge bases in tractable ways, and rich cognitive models that work together with those mechanisms and knowledge bases.

Along with this goes a need for architectures that are likely more heterogeneous. A lot of machine learning to date has focused on relatively homogeneous architectures with individual neurons capable of little more than summation and integration, and often no more than a handful of prespecified modules. As recent work has shown, this is wild oversimplification; at the macro-level, the cortex alone has hundreds of anatomically and likely functionally areas (Van Essen, Donahue, Dierker, & Glasser, 2016); at the micro-level, as mentioned earlier, even a single dendritic compartment of a single neurons can compute the nonlinearity of XOR (Gidon et al., 2020). Adam Marblestone, Tom Dean and I argued (Marcus et al., 2014), the cortex is (contra a common trope) unlikely to compute all of its functions with a single canonical circuit; there is likely to be an important diversity in neural computation that has not yet been captured either in computational neuroscience or in AI.

Two figures capture in a qualitative way what I think we has been going on in recent years, and what we should be going after. The first and most important point of these figures is simply this: the space of potential AI (and machine learning) models is vast, and only a tiny bit of what could exist has been explored. Blank-slate empiricist models have been very well studied, and very well-funded, indulged with computational resources and databases that were unimaginable in the early days of AI; there has been some genuine progress, but brittleness, in so many forms, remains a serious problem; it is time to explore other approaches with similar vigor.

Moving forward requires, minimally, that we build models that in principle can represent and learn the kinds of things that we need for language and higher-level cognition.

Most current systems aren't even in the right ballpark. At a minimum, adequate knowledge frameworks will require that we can represent and manipulate some fraction of our knowledge in algebraic ways, by means of operations over variables; it is likely that some (large) subset of that knowledge is encoded and maintained in terms of structured representations, and much of that knowledge must pertain to and allow the tracking of specific individuals.

Transformer architectures have workarounds for all of this, but in ways that in the end are unlikely to succeed, unless supplemented; at the same time, we absolutely cannot expect that all relevant knowledge is hardwired in advance.

The strong prediction of the current paper is that robust artificial intelligence necessarily will reside in the intersection depicted in Figure 4.

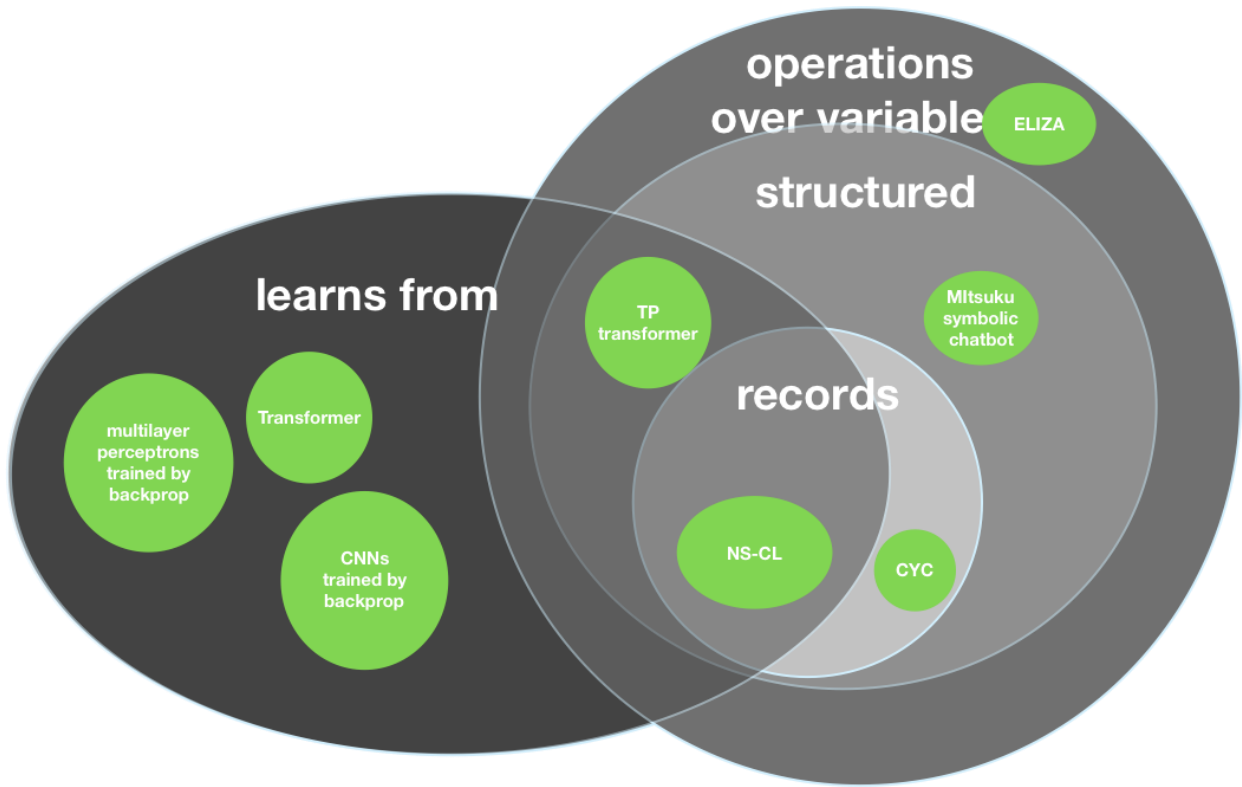

*Figure 4: Venn diagram sketching a few models and architectures within a vast space of possible models of intelligence, focusing on dimensions of learning and symbol-manipulation. The hypothesis of The Algebraic Mind (Marcus, 2001), and core of the present conjecture is that successful models of intelligence will require operation over variables, structured representations, record for individuals. NS-CL [the Neurosymbolic Concept Learner (Mao et al 2019), mentioned in Section 2.1.2] represents one of many possible hybrid models of that sort, many yet to be invented. The thesis of the present article is that this region of intersection should be a central focus of research towards general intelligence in the new decade.*

At the same time, the space of possible models within that intersection is vast, perhaps even infinite; saying that the right architecture is there is a start, but only a start, something like saying that a web browser probably ought to be written in a language that is Turing equivalent. Great, and true, and ... now what? Having the right set of primitives is only a start.

Here's a way to think about this: there are an infinite number of possible computer programs, and only some of them instantiate applications such as (e.g.) web browsers or spreadsheets, and only a subset of them represent web browsers or spreadsheets that are robust. In a similar way, there are infinite number of systems that contain structured representations, records for individuals, operations over variables, all within a framework that allows for learning, but only some of those will instantiate robust intelligences. If the thrust of this article is correct, hybrid architectures that combine learning and symbol manipulation are necessary for robust intelligence, but not sufficient.

One also needs, for example, the right macrostructure, including for instance rich knowledge in multiple domains, as depicted in Figure 5:

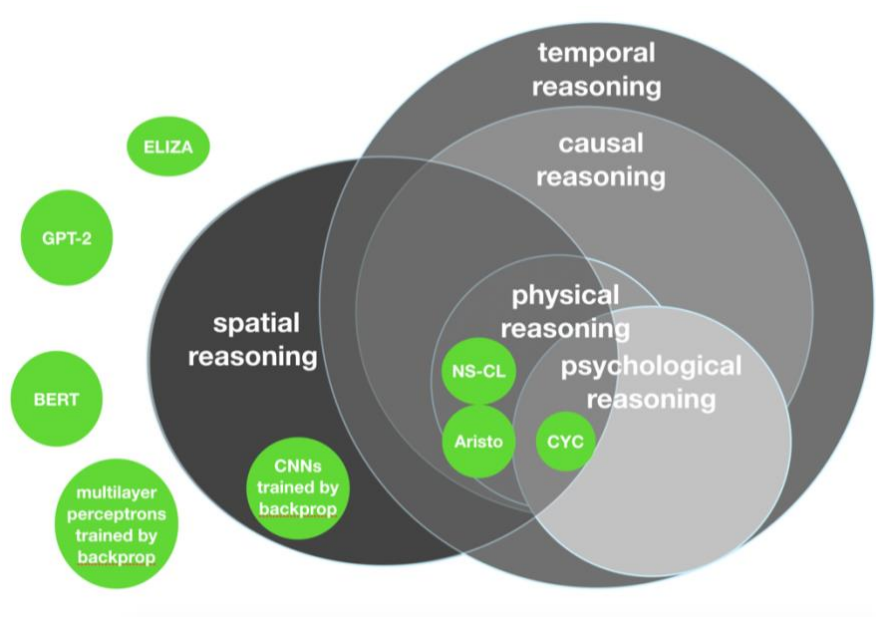

*Figure 5: Venn diagram stressing the need for systems that include machinery for spatial, physical, psychological, temporal, and causal reasoning. Most current neural networks lack explicit mechanisms for these forms of reasoning, and lack natural ways of representing and reasoning over such domains* (but see e.g., Cranmer et al., 2019)

Compare the gist of these two figures with current trends. Most (not quite all) current work in deep learning has eschewed operations over variables, structured representations, and records for individuals; it has similarly typically made do largely without large-scale abstract knowledge, rich cognitive models, and explicit modules for reasoning. There is, by and large, not enough discussion about what the primitives for synthetic cognition need to be. Deep learning has—remarkably—largely achieved what it has achieved without such conventional computational niceties, and without anything that looks like explicit modules for physical reasoning, psychological reasoning and so forth.

But it is a fallacy to suppose that what worked reasonably well for domains such as speech recognition and object labeling—which largely revolve around classification—will necessarily work reliably for language understanding and higher-level reasoning. A number of language benchmarks have been beaten, to be sure, but something profound is still missing. Current deep learning systems can learn endless correlations between arbitrary bits of information, but still go no further; they fail to represent the richness of the world, and lack even any understanding that an external world exists at all.

That's not where we want to be.

Towards the end of Rebooting AI, Ernest Davis and I urged the following

> *In short, our recipe for achieving common sense, and ultimately general intelligence, is this: Start by developing systems that can represent the core frameworks of human knowledge: time, space, causality, basic knowledge of physical objects and their interactions, basic knowledge of humans and their interactions. Embed these in an architecture that can be freely extended to every kind of knowledge, keeping always in mind the central tenets of abstraction, compositionality, and tracking of individuals. Develop powerful reasoning techniques that can deal with knowledge that is complex, uncertain, and incomplete and that can freely work both top-down and bottom-up. Connect these to perception, manipulation, and language. Use these to build rich cognitive models of the world. Then finally the keystone: construct a kind of human-inspired learning system that uses all the knowledge and cognitive abilities that the AI has; that incorporates what it learns into its prior knowledge; and that, like a child, voraciously learns from every possible source of information: interacting with the world, interacting with people, reading, watching videos, even being explicitly taught. Put all that together, and that's how you get to deep understanding.* (Marcus & Davis, 2019).

We concluded "It's a tall order, but it's what has to be done." Even after the dramatic rise of Transformers such GPT-2, which came out after we went to press, I see no reason to change our order.

## 3.2 Is there anything else we can do?

Yes, absolutely.

### 3.2.1 Engineering practice

To begin with, achieving robustness isn't just about developing the right cognitive prerequisites, it is also about developing the right engineering practice. Davis and I discuss this briefly in Chapter X of Rebooting AI, and Tom Dietterich has an excellent discussion in his AAAI Presidential Address (Dietterich, 2017) that I discovered belatedly, after Rebooting AI came out. Davis and I emphasized techniques like redundancy and specifying tolerances that have long served other forms of engineering. Dietterich made eight suggestions, well worth reading, such as constructing optimizations functions to be sensitive to reward and directly constructing machinery for detecting model failures; like us, he also emphasized the need for causal models and the value of redundancy. Joelle Pineau's points about replicability are also essential (Henderson et al., 2017).

*3.2.2. Culture*

There is something else that needs to be fixed, having to do with neither cognitive prerequisites nor sound engineering practice, and that is culture: something is seriously amiss with certain elements of the deep learning community, in a way that is not conducive to progress. This is an elephant in the room, and it must be acknowledged and addressed, if we are to move forward.

In particular outside perspectives, particularly critical ones, are often treated with a kind of extreme aggression (borne of decades of counterproductive hostilities on both side)[19] that should have no place in intellectual discourse, particularly in a field that almost certainly needs to become interdisciplinary if it is to progress.

Students are not blind to this dynamic, and have come to recognize that speaking up for symbol-manipulation as a component to AI can cause damage to their careers. After my debate with Bengio, for example, a young researcher from a prominent deep learning lab wrote to me privately, saying "I've actually wanted to write something … about symbolic AI for two years, and refrained from doing it every time over fear that it could have repercussions of one kind or another on my future career path."

This is a counterproductive state of affairs. As Hinton himself once said, "Max Planck said, 'Science progresses one funeral at a time.'[20] The future depends on some graduate student who is deeply suspicious of everything I have said." Progress often depends on students recognizing the limits of the theories of their elders; if students are afraid to speak, there is a serious problem.

## 3.3. Seeing the whole elephant, a little bit at a time

The good news is that if we can start to work together, progress may not be so far away. If the problem of robust intelligence had already been solved, there would be no need to write this essay at all. But, maybe, just maybe there's enough already out there that if we squint, and look at all the pieces around us, we might be able to imagine what the elephant might look like, if we were to put it all together.

---

[19] A second cultural issue, as one reader of this manuscript pointed out, is that advocates of deep learning have often put far too much stock in big data, often assuming, sometimes incorrectly, that the answers to complex problems can largely be found in ever-bigger data sets and larger and larger clusters of compute. Whole fields, such as linguistics, have largely been dismissed along the way. This cannot be good.

[20] Strictly speaking, Planck never actually said quite that: see
https://quoteinvestigator.com/2017/09/25/progress/

A few thoughts:

- Deep learning has shown us how much can be learned, from massive amounts of data. Co-occurrence statistics and the like may be mere shadows of robust knowledge, but there are sure are a lot of shadows, and maybe we can put those shadows to use, with more sophisticated techniques, so long as we are keenly aware of both their strengths and their limitations.

- CYC shows the potential power of sophisticated reasoning in the presence of rich knowledge bases and rich cognitive models, even if on its own, it is not capable of deriving those models directly from language or perceptual inputs.

- Systems like NS-CL (Mao et al., 2019) show us that symbol manipulation and deep learning can, at least in principle, if not yet at scale, be integrated into a seamless whole that can both perceive and reason.

That's a lot. If we can break out of silos, and cease the hostilities that have slowed progress for six decades, and instead focus on an earnest effort to try bridging these worlds, prospects are good. Mixing metaphors slightly, perhaps the best way to stave off the next possible AI winter may be to rest our tent not on one pole, but on many.

## 3.4 Conclusions, prospects, and implications

Nothing requires us to abandon deep learning, nor ongoing work that focuses on topics such as new hardware, learning rules, evaluation metrics, and training regimes, but it urges a shift from a perspective in which learning is more or less the only first-class citizen to one in which learning is a central member of a broader coalition that is more welcoming to variables, prior knowledge, reasoning, and rich cognitive models.

I have advocated for a four-step program: initial development of hybrid neuro-symbolic architectures, followed by construction of rich, partly-innate cognitive frameworks and large-scale knowledge databases, followed by further development of tools for abstract reasoning over such frameworks, and, ultimately, more sophisticated mechanisms for the representation and induction of cognitive models. Taken together, progress towards these four prerequisites could provide a substrate for richer, more intelligent systems than are currently possible. Ultimately, I think that will redefine what we even mean by *learning*, leading to a (perhaps new) form of learning that traffics in abstract, language-like generalizations, from data, relative to knowledge and cognitive models, incorporating reasoning as part of the learning process.

If none of what I described is individually or even collectively sufficient, it is, I believe, at least enough to bring us much closer to a framework for AI that we can trust.

To put things slightly differently: one approach to research, which is the one I am calling for, would be to identify a well-motivated set of initial primitives (which might include operations over variables, mechanisms for attention, and so forth) first, and then learn about ways of recombining those primitives after, essentially learning what constitutes good practice, given those primitives. Only later, once those principles of good software engineering were settled, might we go on to immensely complex real-world capabilities. Most machine learning work essentially tries to skip the opening steps, tackling complex problems empirically, without ever trying to build a firm understanding about what initial primitives are really required for language and higher-level cognition. Skipping those first steps has not gotten us thus far to language understanding and reliable trustworthy systems that can cope with the unexpected; it is time to reconsider.

In my judgement, we are unlikely to resolve any of our greatest immediate concerns about AI if we don't change course. The current paradigm—long on data, but short on knowledge, reasoning and cognitive models—simply isn't getting us to AI we can trust (Marcus & Davis, 2019). Whether we want to build general purpose robots that live with us in our homes, or autonomous vehicles that drive us around in unpredictable places, or medical diagnosis systems that work as well for rare diseases as for common ones, we need systems that do more than dredge immense datasets for subtler and subtler correlations. In order to do better, and to achieve safety and reliability, we need systems with a rich causal understanding of the world, and that needs to start with an increased focus on how to represent, acquire, and reason with abstract, causal knowledge and detailed internal cognitive models.

§

Rome won't built in a day. Children have a great deal of common sense, can reason, and represent complex knowledge, but it still takes years before they have the sophistication, breadth, and competence of (most) adults. They have started to acquire some of the knowledge, particular aspects of the concrete here and now, but still have to learn, particularly about nuanced domains like politics, economics, sociology, biology, and everyday human interaction.

Figuring out how to reliably build, represent and reason with cognitive models and large-scale background knowledge, presumably by leveraging innovations in hybrid architectures, such as those described in Section 2.1.2, will be an important step, and likely to profitably occupy much of the next decade, but will not be the whole journey.

Importantly, progress in these critical cognitive prerequisites may position AI to be a self-sufficient learner, like a bright school child—but they cannot in themselves provide a guarantee of yielding a complete cognitive being. That said, they might lead to self-teaching machines that are in some ways like a child, with an incomplete understanding of the world but a powerful talent for acquiring new ideas. It's surely just a start, but it will make what has come far seem like mere prelude, to something new that we can't yet fully envision.

## 4. Acknowledgements

In memory of Jacques Mehler, 1936- 2020, scientist, founder of the journal *Cognition*, and great champion of the sort of interdisciplinary cognitive science that we need to take AI to the next level.

This article in part a reflection on the AI Debate I had with Yoshua Bengio on December 23, 2019 in Montreal, Canada, organized by Vince Boucher of Montreal AI. I thank both Yoshua and Vince for making that possible I also thank Deen Abiola, Doug Bemis, Emily Bender, Vince Boucher, Ernie Davis, Tom Dietterich, Pedro Domingos, Chaz Firestone, Artur D'Avila Garcez, Daniel Kahneman, Katia Karpenko, Kristian Kersting, Luis Lamb, Adam Marblestone, Melanie Mitchell, Eyad Nawar, Barney Pell, Jean-Louis Villecroze, and Brad Wyble, who read and commented on early drafts of this manuscript, and Mohamed Amer and Dylan Bourgeois for helpful discussion. Most of all, special thanks go to Ernie Davis, my sounding board for so much in AI; this paper owes a great deal to our conversations, and our joint research.

## 5. References

Adiwardana, D., Luong, M.-T., So, D. R., Hall, J., Fiedel, N., Thoppilan, R. et al. (2020). Towards a Human-like Open-Domain Chatbot. *cs.CL*.

Alcorn, M. A., Li, Q., Gong, Z., Wang, C., Mai, L., Ku, W.-S. et al. (2018). Strike (with) a Pose: Neural Networks Are Easily Fooled by Strange Poses of Familiar Objects. *arXiv*, 1811.11553v3.

Arabshahi, F., Lu, Z., Singh, S., & Anandkumar, A. (2019). Memory Augmented Recursive Neural Networks. *cs.LG*.

Bakken, T. E., Miller, J. A., Ding, S. L., Sunkin, S. M., Smith, K. A., Ng, L. et al. (2016). A comprehensive transcriptional map of primate brain development. *Nature*, *535*(7612), 367-375.

Banino, A., Badia, A. P., Köster, R., Chadwick, M. J., Zambaldi, V., Hassabis, D. et al. (2020). MEMO: A Deep Network for Flexible Combination of Episodic Memories. *cs.LG*.

Bender, E. M., & Lascarides, A. (2019). Linguistic Fundamentals for Natural Language Processing II: 100 Essentials from Semantics and Pragmatics. *Synthesis Lectures on Human Language Technologies*, *12*(3), 1-268.

Bengio, Y. (2019). *From System 1 Deep Learning to System 2 Deep Learning*. Proceedings from NeuripS 2019.

Bengio, Y., Deleu, T., Rahaman, N., Ke, R., Lachapelle, S., Bilaniuk, O. et al. (2019). A Meta-Transfer Objective for Learning to Disentangle Causal Mechanisms. *cs.LG*.

Berent, I., Marcus, G. F., Shimron, J., & Gafos, A. I. (2002). The scope of linguistic generalizations: Evidence from Hebrew word formation. *Cognition*, *83*(2), 113-139.

Berent, I., Vaknin, V., & Marcus, G. F. (2007). Roots, stems, and the universality of lexical representations: Evidence from Hebrew. *Cognition*, *104*(2), 254-286.

Besold, T. R., Garcez, A. D., Stenning, K., van der Torre, L., & van Lambalgen, M. (2017). Reasoning in non-probabilistic uncertainty: Logic programming and neural-symbolic computing as examples. *Minds and Machines*, *27*(1), 37-77.

Bingham, E., Chen, J. P., Jankowiak, M., Obermeyer, F., Pradhan, N., Karaletsos, T. et al. (2019). Pyro: Deep universal probabilistic programming. *The Journal of Machine Learning Research*, *20*(1), 973-978.

Bordes, A., Usunier, N., Chopra, S., & Weston, J. (2015). Large-scale Simple Question Answering with Memory Networks. *arXiv*.

Burgess, C. P., Matthey, L., Watters, N., Kabra, R., Higgins, I., Botvinick, M. et al. (2019). MONet: Unsupervised Scene Decomposition and Representation. *arXiv*, 1901.11390v1.

Carey, S. (2009). *The origin of concepts*. Oxford university press.

Chollet, F. (2019). On the Measure of Intelligence. *cs.AI*.

Clark, P., Etzioni, O., Khashabi, D., Khot, T., Mishra, B. D., Richardson, K. et al. (2019). From 'F' to 'A' on the N.Y. Regents Science Exams: An Overview of the Aristo Project. *cs.CL*.

Cranmer, M. D., Xu, R., Battaglia, P., & Ho, S. (2019). Learning Symbolic Physics with Graph Networks. *arXiv preprint arXiv:1909.05862*.

Cropper, A., Morel, R., & Muggleton, S. (2019). Learning higher-order logic programs. *Machine Learning*, 1-34.

D'Avila Garcez, A. S., Lamb, L. C., & Gabbay, D. M. (2009). *Neural-symbolic cognitive reasoning*. Springer Science & Business Media.

Davis, E. (2019). The Use of Deep Learning for Symbolic Integration: A Review of (Lample and Charton, 2019). *cs.LG*.

Davis, E., Marcus, G., & Frazier-Logue, N. (2017). Commonsense reasoning about containers using radically incomplete information. *Artificial intelligence*, *248*, 46-84.

Dietterich, T. G. (2017). Steps toward robust artificial intelligence. *AI Magazine*, *38*(3), 3-24.

Dyer, F. C., & Dickinson, J. A. (1994). Development of sun compensation by honeybees: how partially experienced bees estimate the sun's course. *Proceedings of the National Academy of Sciences*, *91*(10), 4471-4474.

Engelcke, M., Kosiorek, A. R., Jones, O. P., & Posner, I. (2019). GENESIS: Generative Scene Inference and Sampling with Object-Centric Latent Representations. *arXiv*, 1907.13052v3.

Evans, R., & Grefenstette, E. (2017). Learning Explanatory Rules from Noisy Data. *arXiv*, *cs.NE*.

Fawzi, A., Malinowski, M., Fawzi, H., & Fawzi, O. (2019). Learning dynamic polynomial proofs. *cs.LG*.

Fodor, J. A., & Pylyshyn, Z. W. (1988). Connectionism and cognitive architecture: a critical analysis. *Cognition*, *28*(1-2), 3-71.

Frankland, S. M., & Greene JD. (2019). Concepts and Compositionality: In Search of the Brain's Language of Thought. *Annual review of psychology.*

Gallistel, C. R. (1990). *The organization of learning.* The MIT Press.

Gallistel, C. R., & King, A. P. (2010). *Memory and the computational brain: Why cognitive science will transform neuroscience*. John Wiley & Sons.

George, D., Lehrach, W., Kansky, K., Lázaro-Gredilla, M., Laan, C., Marthi, B. et al. (2017). A generative vision model that trains with high data efficiency and breaks text-based CAPTCHAs. *Science*, *358*(6368).

Gidon, A., Zolnik, T. A., Fidzinski, P., Bolduan, F., Papoutsi, A., Poirazi, P. et al. (2020). Dendritic action potentials and computation in human layer 2/3 cortical neurons. *Science*, *367*(6473), 83-87.

Gopnik, A., & Sobel, D. M. (2000). Detecting blickets: How young children use information about novel causal powers in categorization and induction. *Child development*, *71*(5), 1205-1222.

Goyal, A., Lamb, A., Hoffmann, J., Sodhani, S., Levine, S., Bengio, Y. et al. (2019). Recurrent Independent Mechanisms. *cs.LG*.

Gregor, K., Rezende, D. J., Besse, F., Wu, Y., Merzic, H., & Oord, A. V. D. (2019). Shaping Belief States with Generative Environment Models for RL. *arXiv*, 1906.09237v2.

Gupta, N., Lin, K., Roth, D., Singh, S., & Gardner, M. (2019). Neural Module Networks for Reasoning over Text. *arXiv*, 1912.04971v1.

Ha, D., & Schmidhuber, J. (2018). World Models. *arXiv*, 1803.10122v4.

Henaff, M., Weston, J., Szlam, A., Bordes, A., & LeCun, Y. (2016). Tracking the World State with Recurrent Entity Networks. *arXivICLR 2017*, 1612.03969v3.

Henderson, P., Islam, R., Bachman, P., Pineau, J., Precup, D., & Meger, D. (2017). Deep Reinforcement Learning that Matters. *arXiv*, *cs.LG*.

Hill, F., Lampinen, A., Schneider, R., Clark, S., Botvinick, M., McClelland, J. L. et al. (2019). Emergent systematic generalization in a situated agent. *arXiv preprint arXiv:1910.00571*.

Hinton, G. E. (1990). Preface to the special issue on connectionist symbol processing. *Artificial Intelligence*, *46*(1-2), 1-4.

Janner, M., Levine, S., Freeman, W. T., Tenenbaum, J. B., Finn, C., & Wu, J. (2018). Reasoning About Physical Interactions with Object-Oriented Prediction and Planning. *cs.LG*.

Jia, R., & Liang, P. (2017). Adversarial Examples for Evaluating Reading Comprehension Systems. *arXiv*.

Johnson-Laird, P. N. (1983). *Mental models: Towards a cognitive science of language, inference, and consciousness* ((6)). Harvard University Press.

Kansky, K., Silver, T., Mély, D. A., Eldawy, M., Lázaro-Gredilla, M., Lou, X. et al. (2017). Schema Networks: Zero-shot Transfer with a Generative Causal Model of Intuitive Physics. *arXIv*, *cs.AI*.

Keil, F. C. (1992). *Concepts, kinds, and cognitive development*. mit Press.

Knudsen, E. I., & Konishi, M. (1979). Mechanisms of sound localization in the barn owl (Tyto alba). *Journal of Comparative Physiology*, *133*(1), 13-21.

Koul, A., Greydanus, S., Fern - arXiv preprint arXiv:1811.12530, A., & 2018. Learning finite state representations of recurrent policy networks. *arxiv.org*.

Lake, B. M., & Baroni, M. (2017). Still not systematic after all these years: On the compositional skills of sequence-to-sequence recurrent networks. *arXiv*.

Lample, G., & Charton, F. (2019). Deep Learning for Symbolic Mathematics. *arXiv*, 1912.01412v1.

Landau, B., Gleitman, L. R., & Landau, B. (2009). *Language and experience: Evidence from the blind child* (8). Harvard University Press.

LeCun, Y. (1989). Generalization and network design strategies. *Technical Report CRG-TR-89-4*.

Legenstein, R., Papadimitriou, C. H., Vempala, S., & Maass, W. (2016). Assembly pointers for variable binding in networks of spiking neurons. *arXiv*, 1611.03698v1.

Lenat, D. (2019). What AI Can Learn From Romeo & Juliet. *Forbes*.

Lenat, D. B., Prakash, M., & Shepherd, M. (1985). CYC: Using common sense knowledge to overcome brittleness and knowledge acquisition bottlenecks. *AI magazine*, *6*(4), 65-65.

Leslie, A. M. (1982). The perception of causality in infants. *Perception*, *11*(2), 173-186.

Maier, A., Schebesch, F., Syben, C., Würfl, T., Steidl, S., Choi, J.-H. et al. (2017). Precision Learning: Towards Use of Known Operators in Neural Networks. *A. Maier, F. Schebesch, C. Syben, T. W\"urfl, S. Steidl, J.-H. Choi, R. Fahrig, Precision Learning: Towards Use of Known Operators in Neural Networks, in: 24rd International Conference on Pattern Recognition (ICPR), 2018, pp. 183-188*, *cs.CV*.

Mandler, J. M. (1992). How to build a baby: II. Conceptual primitives. *Psychological review*, *99*(4), 587.

Mao, J., Gan, C., Kohli, P., Tenenbaum, J. B., & Wu, J. (2019). The Neuro-Symbolic concept learner: Interpreting scenes, words, and sentences from natural supervision. *arXiv preprint arXiv:1904.12584*.

Marcus, G. (2019). *Deep Understanding: The Next Challenge for AI*. Proceedings from NeurIPS 2019.

Marcus, G. (2020). GPT-2 and the Nature of Intelligence. *The Gradient*.

Marcus, G., Marblestone, A., & Dean, T. (2014). The atoms of neural computation. *Science*, *346*(6209), 551-552.

Marcus, G. (2018). Deep Learning: A Critical Appraisal. *arXiv*.

Marcus, G., & Davis, E. (2019). *Rebooting AI: building artificial intelligence we can trust*. Pantheon.

Marcus, G. F. (2008). *Kluge : the haphazard construction of the human mind*. Boston: Houghton Mifflin.

Marcus, G. F. (2001). *The Algebraic Mind: Integrating Connectionism and cognitive science*. Cambridge, Mass.: MIT Press.

Marcus, G. F. (2004). *The Birth of the Mind : how a tiny number of genes creates the complexities of human thought*. Basic Books.

Marcus, G. F. (1998). Rethinking eliminative connectionism. *Cogn Psychol*, *37*(3), 243-282.

Marcus, G. F., Pinker, S., Ullman, M., Hollander, M., Rosen, T. J., & Xu, F. (1992). Overregularization in language acquisition. *Monogr Soc Res Child Dev*, *57*(4), 1-182.

Marcus, G. F., Vijayan, S., Bandi Rao, S., & Vishton, P. M. (1999). Rule learning by seven-month-old infants. *Science*, *283*(5398), 77-80.

Marr, D. (1982). *Vision: A Computational Investigation into the Human Representation and Processing of Visual Information*. San Francisco: WH Freeman and Co.

McClelland, J. L. (2019). Integrating New Knowledge into a Neural Network without Catastrophic Interference: Computational and Theoretical Investigations in a Hierarchically Structured Environment.

McCloskey, M., & Cohen, N. J. (1989). Catastrophic interference in connectionist networks: The sequential learning problem. *Elsevier*, *24*, 109-165.

Miller, J. A., Ding, S. L., Sunkin, S. M., Smith, K. A., Ng, L., Szafer, A. et al. (2014). Transcriptional landscape of the prenatal human brain. *Nature*, *508*(7495), 199-206.

Minervini, P., Bošnjak, M., Rocktäschel, T., Riedel, S., & Grefenstette, E. (2019). Differentiable Reasoning on Large Knowledge Bases and Natural Language. *cs.LG*.

Mnih, V., Kavukcuoglu, K., Silver, D., Rusu, A. A., Veness, J., Bellemare, M. G. et al. (2015). Human-level control through deep reinforcement learning. *Nature*, *518*(7540), 529-533.

Newell, A. (1980). Physical symbol systems. *Cognitive science*, *4*(2), 135-183.

Nie, Y., Williams, A., Dinan, E., Bansal, M., Weston, J., & Kiela, D. (2019). Adversarial NLI: A New Benchmark for Natural Language Understanding. *cs.CL*.

Norvig, P. (1986). Unified theory of inference for text understanding.

OpenAI, Akkaya, I., Andrychowicz, M., Chociej, M., Litwin, M., McGrew, B. et al. (2019). Solving Rubik's Cube with a Robot Hand. *arXiv*, 1910.07113v1.

Pasupat, P., & Liang, P. (2015). Compositional semantic parsing on semi-structured tables. *arXiv preprint arXiv:1508.00305*.

Pearl, J., & Mackenzie, D. (2018). *The book of why: the new science of cause and effect*. Basic Books.

Polozov, O., & Gulwani, S. (2015). *FlashMeta: a framework for inductive program synthesis*.

Rabinowitz, N. C., Perbet, F., Song, H. F., Zhang, C., Eslami, S. M. A., & Botvinick, M. (2018). Machine Theory of Mind. *arXiv*, 1802.07740v2.

Radford, A., Wu, J., Child, R., Luan, D., Amodei, D., & Sutskever, I. (2019). Language models are unsupervised multitask learners. *OpenAI Blog*, *1*(8).

Raedt, L. D., Kersting, K., Natarajan, S., & Poole, D. (2016). Statistical relational artificial intelligence: Logic, probability, and computation. *Synthesis Lectures on Artificial Intelligence and Machine Learning*, *10*(2), 1-189.

Richardson, M., & Domingos, P. (2006). Markov logic networks. *Machine learning*, *62*(1), 107-136.

Schank, R. C., & Abelson, R. P. (1977). *Scripts, plans, goals, and understanding: An inquiry into human knowledge structures*. Erlbaum.

Schlag, I., Smolensky, P., Fernandez, R., Jojic, N., Schmidhuber, J., & Gao, J. (2019). Enhancing the Transformer with Explicit Relational Encoding for Math Problem Solving. *cs.LG*.

Schrittwieser, J., Antonoglou, I., Hubert, T., Simonyan, K., Sifre, L., Schmitt, S. et al. (2019). Mastering Atari, Go, Chess and Shogi by Planning with a Learned Model. *arXiv*, 1911.08265v1.

Serafini, L., & Garcez, A. D. (2016). Logic Tensor Networks: Deep Learning and Logical Reasoning from Data and Knowledge. *arXiv*, 1606.04422v2.

Shavlik, J. W. (1994). Combining symbolic and neural learning. *Machine Learning*, *14*(3), 321-331.

Silver, D., Schrittwieser, J., Simonyan, K., Antonoglou, I., Huang, A., Guez, A. et al. (2017). Mastering the game of Go without human knowledge. *Nature*, *550*(7676), 354-359.

Smolensky, P., Lee, M., He, X., Yih, W.-t., Gao, J., & Deng, L. (2016). Basic Reasoning with Tensor Product Representations. *arXiv*, *cs.AI*.

Spelke, E. (1994). Initial knowledge: six suggestions. *Cognition*, *50*(1-3), 431-445.

Sun, R. (1996). Hybrid Connectionist-Symbolic Modules: A Report from the IJCAI-95 Workshop on Connectionist-Symbolic Integration. *AI Magazine*, *17*(2), 99-99.

Tanenhaus, M. K., Spivey-Knowlton, M. J., Eberhard, K. M., & Sedivy, J. C. (1995). Integration of visual and linguistic information in spoken language comprehension. *Science*, *268*(5217), 1632-1634.

Van den Broeck, G. (2019). *IJCAI-19 Computers and Thought Award Lecture*. Proceedings from IJCAI-19.

Van Essen, D. C., Donahue, C., Dierker, D. L., & Glasser, M. F. (2016). Parcellations and connectivity patterns in human and macaque cerebral cortex. In *Micro-, Meso-and Macro-Connectomics of the Brain* (pp. 89-106). Springer, Cham.

Vaswani, A., Shazeer, N., Parmar, N., Uszkoreit, J., Jones, L., Gomez, A. N. et al. (2017). Attention Is All You Need. *cs.CL*.

Veerapaneni, R., Co-Reyes, J. D., Chang, M., Janner, M., Finn, C., Wu, J. et al. (2019). Entity Abstraction in Visual Model-Based Reinforcement Learning. *cs.LG*.

Vergari, A., Di Mauro, N., & Van den Broek, G. (2019). *Tutorial slides on tractable probabilistic models*.

Wayne, G., Hung, C.-C., Amos, D., Mirza, M., Ahuja, A., Grabska-Barwinska, A. et al. (2018). Unsupervised Predictive Memory in a Goal-Directed Agent. *arXiv*, 1803.10760v1.

Winograd, T. (1971). Procedures as a representation for data in a computer program for understanding natural language.

Yang, F., Yang, Z., & Cohen, W. W. (2017). Differentiable Learning of Logical Rules for Knowledge Base Reasoning. *cs.AI*.

Zhang, R., Wu, J., Zhang, C., Freeman, W. T., & Tenenbaum, J. B. (2016). A Comparative Evaluation of Approximate Probabilistic Simulation and Deep Neural Networks as Accounts of Human Physical Scene Understanding. *arXiv*, 1605.01138v2.